\documentclass{article}

\usepackage{microtype}
\usepackage{graphicx}
\usepackage{subfigure}
\usepackage{booktabs} 

\usepackage{amsthm} 
\usepackage{enumitem}
\usepackage{bbm}
\usepackage{amsmath,amssymb,amsfonts}
\usepackage{bm}
\usepackage{eqparbox}
\usepackage{graphicx}
\usepackage{xspace}
\usepackage{comment}
\usepackage{multirow}
\usepackage{verbatim}
\usepackage{arydshln}
\usepackage{tabularx}
\usepackage{tabulary}
\usepackage{booktabs}
\usepackage{makecell}
\usepackage{xspace}
\usepackage{circledsteps}
\usepackage{textcomp}
\usepackage{wasysym}

\usepackage{wrapfig}

\usepackage[T1]{fontenc}

\usepackage[hang,flushmargin]{footmisc}
\theoremstyle{definition}
\newtheorem{definition}{Definition}

\newcommand{\cmmnt}[1]{\ignorespaces} 

\newcommand{\Hquad}{\hspace{0.5em}}

\newcommand{\algname}{{AdaPromptCL}}

\makeatletter
\newcommand*\bigcdot{\mathpalette\bigcdot@{.5}}
\newcommand*\bigcdot@[2]{\mathbin{\vcenter{\hbox{\scalebox{#2}{$\m@th#1\bullet$}}}}}
\makeatother

\usepackage{hyperref}

\usepackage[accepted]{icml2024}

\usepackage{amsmath}
\usepackage{amssymb}
\usepackage{mathtools}
\usepackage{amsthm}

\usepackage{balance}

\usepackage{fix-cm}
\usepackage[T1]{fontenc}

\usepackage[capitalize,noabbrev]{cleveref}

\theoremstyle{plain}

\theoremstyle{remark}

\icmltitlerunning{One Size Fits All for Semantic Shifts: Adaptive Prompt Tuning for Continual Learning}

\begin{document}

\twocolumn[
\icmltitle{One Size Fits All for Semantic Shifts: \\ Adaptive Prompt Tuning for Continual Learning}



\icmlsetsymbol{equal}{*}

\begin{icmlauthorlist}
\icmlauthor{Doyoung Kim}{sch}
\icmlauthor{Susik Yoon}{sch2}
\icmlauthor{Dongmin Park}{comp}
\icmlauthor{Youngjun Lee}{sch}
\icmlauthor{Hwanjun Song}{sch}
\icmlauthor{Jihwan Bang}{sch}
\icmlauthor{Jae-Gil Lee}{sch}
\end{icmlauthorlist}

\icmlaffiliation{comp}{KRAFTON, Seoul, Republic of Korea}
\icmlaffiliation{sch}{KAIST, Daejeon, Republic of Korea}
\icmlaffiliation{sch2}{Korea University, Seoul, Republic of Korea}

\icmlcorrespondingauthor{Jae-Gil Lee}{jaegil@kaist.ac.kr}

\icmlkeywords{Machine Learning, ICML}

\vskip 0.3in
]



\printAffiliationsAndNotice{}  

\begin{abstract}
In real-world continual learning\,(CL) scenarios, tasks often exhibit intricate and unpredictable semantic shifts, posing challenges for \textit{fixed} prompt management strategies which are tailored to only handle semantic shifts of \emph{uniform} degree (i.e., uniformly mild or uniformly abrupt).  
To address this limitation, we propose an \emph{adaptive} prompting approach that effectively accommodates semantic shifts of \emph{varying} degree where mild and abrupt shifts are mixed. \algname{} employs the assign-and-refine semantic grouping mechanism that dynamically manages prompt groups in accordance with the semantic similarity between tasks, enhancing the quality of grouping through continuous refinement. Our experiment results demonstrate that \algname{} outperforms existing prompting methods by up to 21.3\%, especially in the benchmark datasets with diverse semantic shifts between tasks.


\end{abstract}

\section{Introduction}
\label{sec:intro}


\begin{figure*}[!t]
    \begin{center}
    \includegraphics[width=0.99\textwidth]{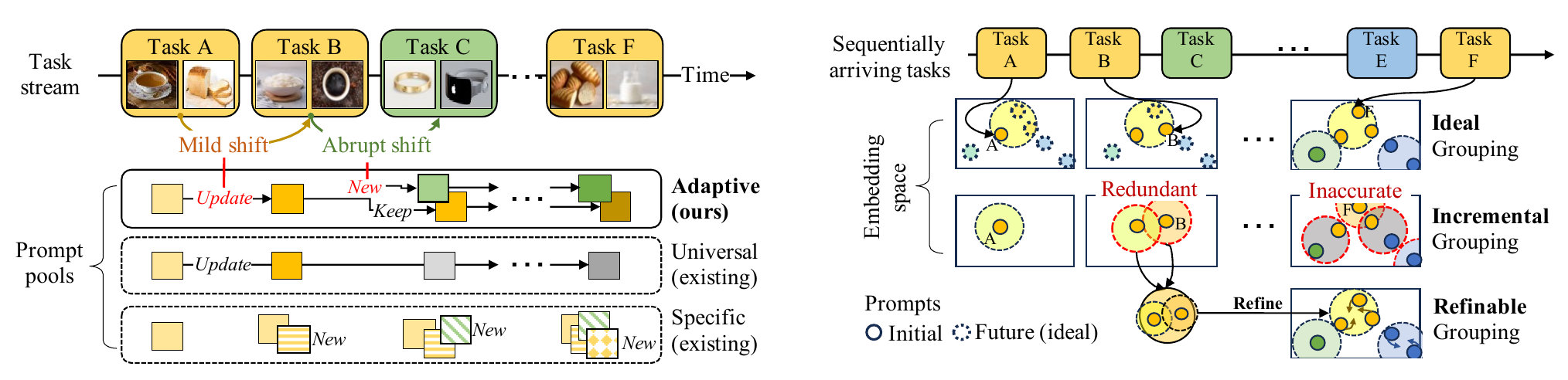}
        \end{center}
    \vspace*{-0.0cm}
    \hspace*{2cm} {\small (a) Comparison of prompting strategies.} \hspace*{3.5cm} {\small (b) Challenge in semantic grouping.}
    \vspace*{-0.2cm}
    \caption{\textbf{Key idea of \algname{}}: (a) shows the comparison of our adaptive prompting against existing prompting CL methods on a practical CL scenario; (b) depicts the inherent challenges in implementing adaptive prompting. Colors indicate task semantics. 
    }
    \label{fig:overview}
    \vspace*{-0.2cm}
\end{figure*}

Recently, rehersal-free continual learning\,(CL) methods \cite{wang2022learning, wang2022sprompts, wang2022dualprompt, gao2023unified} have gained interest because of their superior performance even without using previous samples. Small learnable parameters, known as \textit{prompts}, are added to input data of various tasks for refining a pre-trained model. These methods can be further categorized into \textit{universal} and \textit{specific} prompting methods, depending on how the prompts are managed for sequentially arriving tasks. Universal prompting methods\,(e.g., LAE\,\cite{gao2023unified}) train a fixed set of prompts consistently used for all tasks. This strategy captures generalizable knowledge accumulated from all prior tasks which is believed to be beneficial for upcoming similar tasks. On the other hand, specific prompting methods\,(e.g., S-Prompts\,\cite{wang2022sprompts}) train prompts individually designated for each task, mainly to alleviate catastrophic forgetting which becomes more pronounced with highly divergent tasks.

\smallskip
While these two approaches have proven effective in their respective CL scenarios---similar tasks for universal prompting and distinct tasks for specific prompting, real-world scenarios frequently involve intricate and unpredictable semantic shifts between tasks over time. Let us specify the \emph{degree} of semantic shifts by considering whether a higher-level category of samples undergoes a change. For example, in Figure \ref{fig:overview}(a), the shift from Task A to Task B is \emph{mild} because both tasks involve product classification in the food and beverage category; on the other hand, the shift from Task B to Task C is \emph{abrupt} because Task C belongs to the apparel and accessories category. This example illustrates a combination of \emph{semantic shifts of varying degrees}.

\smallskip
Existing prompting methods, however, are not suitable to accommodate semantic shifts of varying degrees, but can only accommodate those of \emph{uniform} degree. In detail, universal prompting supports uniformly mild shifts, while specific prompting supports uniformly abrupt shifts. On the contrary, for a combination of mild and abrupt shifts in Figure \ref{fig:overview}(a), universal prompting incurs insufficient\,(one instead of two) prompts, and specific prompting incurs redundant\,(four instead of two) prompts. Consequently, the existing methods reach a sub-optimal solution, compromising between forgetting prevention and knowledge transfer.

\smallskip
In this paper, we contend that prompting methods should accommodate semantic shifts of \emph{varying} degrees. As a consequence, we propose an \emph{adaptive} prompting strategy to address the limited effectiveness of the \emph{fixed} prompting strategy---either universal or specific. By carefully considering task semantics, adaptive prompting achieves minimal yet adequate prompts that effectively accommodate varying semantic shifts between tasks. As shown in Figure \ref{fig:overview}(a), it appropriately updates an existing prompt for mildly shifted tasks\,(e.g., Task B) and introduces a new prompt for abruptly shifted tasks\,(e.g., Task C).

\smallskip
The key issue here is grouping semantically similar tasks so as to \emph{share} a prompt among those tasks. This grouping is especially challenging because the number and nature of tasks are \emph{not} predetermined in a real-world CL setting. It is widely known that the quality of online clustering\,(e.g., BIRCH) can be influenced by the order of data insertion\,\cite{zhang1996birch}. As shown in Figure \ref{fig:overview}(b), with prior knowledge on future tasks\,(see ``Ideal Grouping''), a single optimal prompt can be easily derived for the three tasks in yellow. In practice, as tasks arrive sequentially\,(see ``Incremental Grouping''), na\"ive incremental grouping may inadvertently create \emph{redundant} prompts that separately cover similar tasks; even worse, these redundant prompts result in \emph{inaccurate} prompts that contain the tasks of different semantics.

\smallskip
Our prompt-based CL framework, named \algname{}, incorporates an innovative mechanism of \emph{assign-and-refine} semantic grouping in order to realize adaptive prompting. Because the future tasks are unknown as aforementioned, in the assignment step, each new task is added to one of existing groups or a new group based on its similarity to the existing groups. Importantly, to prepare potential refinement of groups, potentially possible groups are stored as well. In the refinement step, if an incoming task (e.g., Task F in Figure \ref{fig:overview}(b)) provides a clear clue to a better semantic grouping, the existing groups are refined accordingly, and their prompts are retrieved from those maintained for potential groups (see ``Refinable Grouping'').

\begin{figure}[t!]
\centering 
\includegraphics[width=0.87\columnwidth]{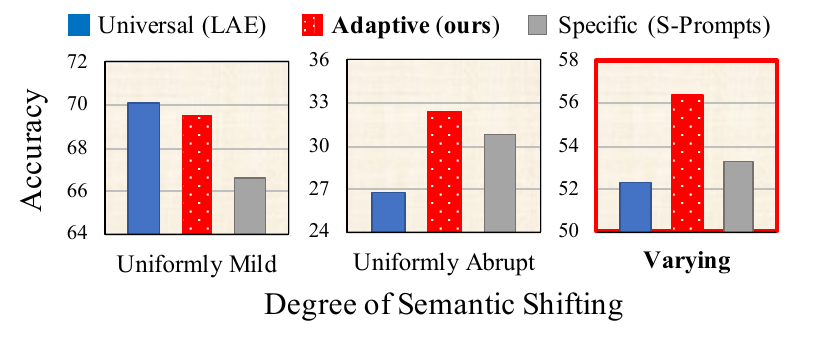}
\vspace*{-0.3cm}
\caption{Performance of representative methods for different prompting strategies across various degrees of semantic shifts.}
\vspace*{-0.4cm}
\label{fig:perf_overview}
\end{figure}

\smallskip
To the best of our knowledge, this is the first work to examine the effectiveness of prompt tuning-based methods in CL across a spectrum of task semantic shifts. In Section \ref{sec:evaluation}, a thorough assessment is performed by categorizing CL scenarios into uniform and varying semantic shifts. As briefed in Figure \ref{fig:perf_overview}, \algname{} demonstrates consistently outstanding performance across both scenarios without regard to the degree of semantic shifts, whereas the fixed prompting methods achieve satisfactory results only within their respective scenarios. In particular, when applied to a realistic CL scenario with varying semantic shifts, \algname{} significantly surpasses the existing methods by up to 13.6\% in terms of accuracy.

\section{Related Work}
\label{sec:related_works}

\subsection{Continual Learning}
\label{subsec:continual learning}

Continual learning\,(CL) confronts the stability-plasticity dilemma inherent in learning a sequence of tasks. This dilemma describes the challenge of preserving acquired knowledge\,(stability) while incorporating new information\,(plasticity). 
Representative CL approaches can be categorized into \textit{rehearsal-based} methods, which rely on retaining and replaying subsets of past data\,\cite{rolnick2019experience, bang2021rainbow, buzzega2020dark, aljundi2019mir, shim2021online, aljundi2019gradient, prabhu2020gdumb, koh2022online, yoon2023scstory, kim2024adaptive}; \textit{regularization} techniques, which constrain updates to preserve past knowledge\,\cite{chaudhry2018riemannian, kirkpatrick2017overcoming}; and \textit{architectural} solutions that reconfigure the model structure for new tasks\,\cite{mallya2018packnet, aljundi2017expert, lee2020neural}. These strategies are detailed in several comprehensive surveys\,\cite{de2021continual, mai2022online, mundt2023wholistic}.
\vspace{-0.2cm}

\subsection{Prompt-based Continual Learning}
\label{subsec:pt continual learning}

The emergence of \textit{rehearsal-free} methods has offered a novel perspective on CL by exploiting the potential of pre-trained models, such as ViT\,\cite{dosovitskiy2021an}, which have demonstrated a remarkable ability to grasp general representation. The rehearsal-free methods perform fine-tuning through prompts, i.e., small, learnable weight parameters that refine the model's representation. This approach brings significant memory and computational efficiency gains since it only modifies a small part of the model's parameters for each new task. 

The rehearsal-free strategies are divided into \textit{universal} and \textit{specific} prompting methods, as aforementioned. In universal prompting, VPT\,\cite{jia2022visual} and L2P\,\cite{wang2022learning} respectively optimize a single prompt and a shared pool of prompts; LAE\,\cite{gao2023unified} advances them by accumulating and ensembling prompts over time. In specific prompting, S-Prompts\,\cite{wang2022sprompts} aims to train prompts to individual tasks to address catastrophic forgetting. Meanwhile, DP\,\cite{wang2022dualprompt} employs the simultaneous use of both universal and specific prompts for all tasks. However, most rehearsal-free methods have not adequately considered the semantic relationships between tasks, which could lead to suboptimal outcomes due to the lack of shared learning across semantically related tasks. 
This paper addresses this gap by integrating task semantics into prompt-tuning strategies.
\vspace{-0.1cm}

\subsection{Semantic Grouping for Multi-Task Learning}
\label{subsec:Semantic-based grouping}

In multi-task learning\,(MTL), the ability to effectively share knowledge across diverse tasks is crucial for enhancing problem-solving capabilities\,\cite{guo2020learning, wu2020understanding, liu2023hipro, yang2023adatask}. However, arbitrarily combining tasks can result in conflicting task groups in which the tasks interfere with one another, resulting in a degradation in performance, a phenomenon known as negative transfer\,\cite{wu2020understanding}. 
Addressing this issue, \textit{semantic-based task grouping} methods\,\cite{zamir2018taskonomy, standley2020tasks, fifty2021efficiently, song2022efficient, liu2023hipro} have been introduced, which emphasize clustering semantically similar tasks to amplify positive transfer and boost performance.
An intrinsic need for specific models for distinct task groups has driven the adoption of parameter-efficient techniques such as prompt tuning\,\cite{liu2023hipro}.
However, their direct adoption poses challenges because they assume simultaneous availability of all tasks, contradicting practical CL environments. 
\vspace{-0.1cm}
\section{Preliminaries}
\label{sec:preliminaries}

\subsection{Continual Learning}

We consider a sequence of tasks $\mathcal{T}=\langle \tau^1, \dots, \tau^{|\mathcal{T}|} \rangle$ for a data stream $\mathcal{D} = \langle D^{1}, \dots, D^{|\mathcal{T}|}\rangle$, where each sub-dataset $D_{t} = \{(x_i^t, y_i^t)\}_{i}$ for the $t$-th task is derived from a joint distribution of an input space $\mathcal{X}^{t}$ and a label space $\mathcal{Y}^{t}$. The goal of CL is to continuously learn a classifier that maximizes test accuracy for all encountered tasks $\{\tau^1,\tau^2,\dots,\tau^t\}$, without having access to data streams $D^{t^\prime<t}$ of prior tasks. These data streams may come from diverse environments, and thus the associated tasks exhibit semantic shifts of varying degrees, calling for adaptive CL approaches.

\subsection{Prompt-based Continual Learning}

\noindent \textbf{Definition of a Prompt.}
Recent prompt-based CL methods\,\cite{wang2022dualprompt, wang2022learning, gao2023unified} mostly employ a pretrained backbone (e.g., ViT\,\cite{dosovitskiy2021an}) with $L$ transformer blocks, where each of these blocks utilizes a multi-head attention module to extract pertinent features. A \emph{prompt} refers to a set of vectors that provide auxiliary context or guide the attention mechanism towards specific patterns in the data for a task,
\begin{equation}
{\rm P} = [{\rm \mathbf{p}}_1,\dots, {\rm \mathbf{p}}_p ] \in \mathbb{R}^{p \times d},
\label{eq:prompt}
\end{equation}
where $p$ and $d$ are the number and the dimensionality of tokens ${\rm \mathbf{p}}_i$, respectively. The prompt serves as a prefix to the input for the transformer block's attention module\,\cite{wang2022dualprompt, gao2023unified}.

\noindent \textbf{Prompt Tuning and Inference.}
Universal and specific strategies are used for managing and selecting prompts. 
First, \emph{universal} prompting employs a constant prompt ${\rm P}$ on all tasks. The optimization objective for a given task $\tau^t$ with $D^t$ is to minimize the cross-entropy loss $\ell_{ce}$,
\begin{equation}
\small
\min_{{\rm P}, {\rm k}, \phi} \ell_{ce} (f_\phi(f([x^t; {\rm P}])), y^t), 
\label{eq:objective}
\end{equation}
where $f_\phi$ is a classifier, and $f$ is a pre-trained model that receives the concatenation of the input tokens $x^t$ and the prompt ${\rm P}$ and then returns the $[\mathtt{CLS}]$ token for prediction.
Second, \emph{specific} prompting employs a collection of task-specific prompts $\{{\rm P}^t \mid \tau^t\in\mathcal{T} \}$ and learnable prompt keys $\{{\rm k}^t \mid \tau^t\in\mathcal{T} \}$ to determine the prompt to be used. The optimization objective is to minimize the distance between the input and the prompt key, in addition to minimizing the cross-entropy loss,
\begin{equation}
\small
\min_{{\rm P}^t, {\rm k}^t, \phi} \ell_{ce} (f_\phi(f([x^t; {\rm P}^t])), y^t) + \lambda{\rm d}(f(x^t),{\rm k}^t),
\label{eq:spcf_objective}
\end{equation}
where $\lambda$ denotes a balancing coefficient and $\rm{d}(\cdot,\cdot)$ represents a distance function.

During inference, given a testing instance $x$, the matching prompt ${\rm P}^{\hat{t}}$ is chosen to minimize the distance between $f(x^t)$ and its prompt key ${\rm k}^{t}$,
\begin{equation}
\hat{t} = \arg\min_{t} {\rm d}(f(x^t), {\rm k}^{t}).
\label{eq:preliminary-inference}
\end{equation}


\section{\algname{}: Adaptive Prompting via Semantic Grouping}
\label{sec:methodology}

\subsection{Problem Statement}
\label{subsec:problem}

Our \emph{adaptive} prompting is in between the two extremes---universal and specific prompting. That is, at the $t$-th task, while universal and specific prompting strategies maintain \emph{one} and $t$ prompts, respectively, our adaptive prompting strategy manages \emph{from one to $t$} prompts depending on the semantic shifts observed so far. Thus, the key challenge is to lie in the management of the best collection of prompts, which is realized by our novel \emph{semantic grouping} process. 
A prompt is assigned to \textrm{each semantic group}, with a variable number of prompts ranging from \emph{one} to $t$.
Specifically, for a sequence of CL tasks $\mathcal{T}$, this grouping process maintains $\mathcal{G}=\{G_1,\dots, G_{|\mathcal{G}|}\}$ that minimizes
\begin{equation}
J(\mathcal{G}) =  \sum_{G \in \mathcal{G}} \sum_{\tau^i, \tau^j \in G} {\rm d}\big( {\rm s}(\tau^i), {\rm s}(\tau^j)\big) + \alpha|\mathcal{G}|,
\label{eq:obj}
\end{equation}
where $G$ is a group of tasks and ${\rm s}(\tau)$ is a semantic representation of a task $\tau$ which can be derived by a task-specific prompt ${\rm P}_\tau$\footnote{Task-specific prompts are known to capture the inherent patterns within the distributions of task data. See Definition \ref{def:task_semantic_embedding} for details.}. 
$\alpha|\mathcal{G}|$ penalizes an excessive number of groups so that they capture the general knowledge from the tasks. A group-wise prompt ${\rm P}_{G}$ is assigned to each group $G$.

Finding optimal semantic groups $\mathcal{G}^*$ for all tasks $\mathcal{T}$ through optimization of Eq.~\eqref{eq:obj} is infeasible in CL, as the data streams of future tasks $D^{t^\prime>t}$ are unknown in practice. A straightforward solution is an incremental greedy approach, where each incoming task is optimally assigned to an existing group or initiated as a new group, depending on its semantic similarity to the current set of groups. However, as demonstrated in Figure \ref{fig:overview}(b), this method may result in a local-optimal grouping. \emph{Refining} this local-optimal grouping whenever possible is essential to attain the global-optimal grouping for all tasks. 


\subsection{Overview of \algname{}}
\label{subsec:overview}


\setlength{\columnsep}{0.2cm}%
\begin{wrapfigure}{r}{0.22\textwidth}
    \vspace*{-0.2cm}
    \includegraphics[width=0.22\textwidth, height=5.7cm]{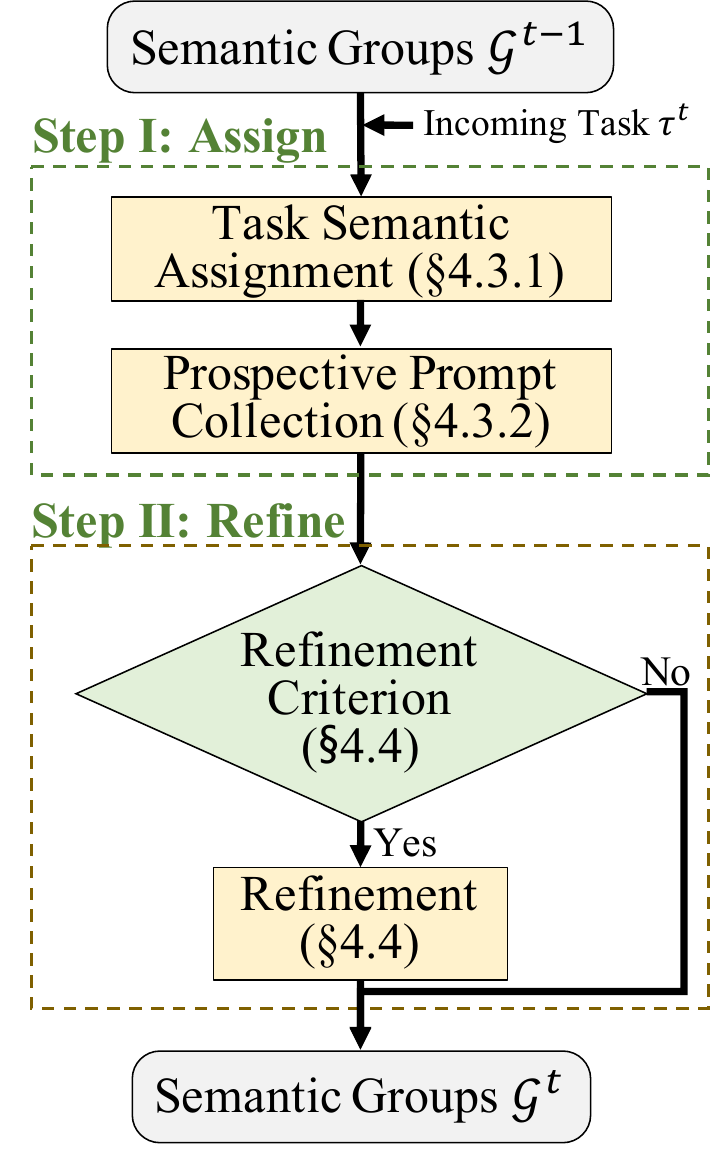} \\
    \vspace*{-0.6cm}
    \caption{Overall flow.}
    \label{fig:flow}
    \vspace*{-0.3cm}
\end{wrapfigure}

\begin{figure*}[t!]
\centering 
\includegraphics[width=0.85\textwidth]{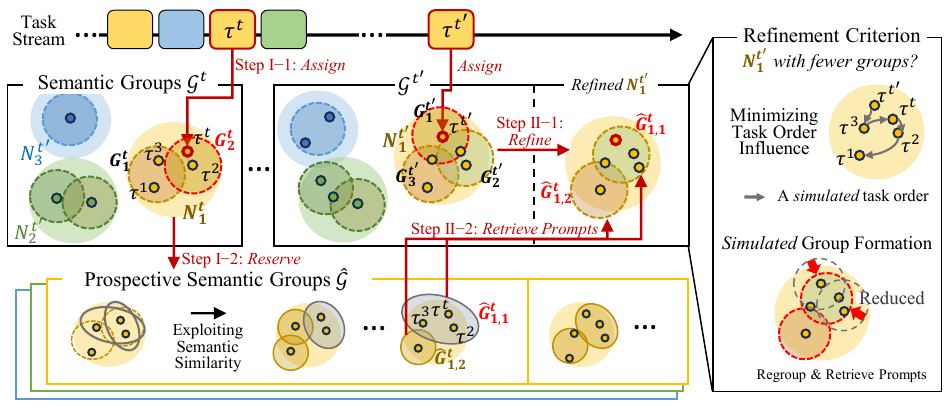}
\vspace*{-0.1cm}
\caption{{Detailed view of the assign-and-refine semantic grouping process.} 
{\textbf{I}.\ The assignment step adds the task $\tau^t$ to the group $G_2^t$ and reserves a prospective group $\hat{G}^t_{1,1}$ with its prompt. \textbf{II}.\ The refinement step, assuming that a refinement is needed when the task $\tau^{t^\prime}$ is received, three groups $G^{t^\prime}_1$, $G^{t^\prime}_2$, and $G^{t^\prime}_3$ are reduced to fewer prospective groups $\hat{G}^t_{1,1}$ and $\hat{G}^t_{1,2}$, with their prompts retrieved.}
}
\vspace*{-0.1cm}
\label{fig:Outline}
\end{figure*}

Figure \ref{fig:flow} illustrates the \textit{assign-and-refine} semantic grouping framework of \algname{}. First, the \emph{assignment} step adds an incoming task $\tau^t$ to an appropriate semantic group and prepares potential refinement by proactively reserving \emph{prospective} semantic groups and their prompts. Second, the \emph{refinement} step improves the quality of semantic grouping if possible and reflects a refinement by retrieving the prospective prompts. 
Figure \ref{fig:Outline} as well as Appendix \ref{app_sec:overall_pseudocode} detail each step.

\subsection{Step I: Semantic Assignment}
\label{subsec:mac}

\subsubsection{Task Semantic Assignment}
For an incoming task $\tau^t$, we form a set of \emph{semantic groups} ${\mathcal{G}}^{t}=\{G^t_1, \cdots, G^t_{|{\mathcal{G}}^{t}|}\}$ in Definition \ref{def:macro_semantic_group}.

\begin{definition}{\sc (Semantic Group)}
\label{def:macro_semantic_group}
A \textit{semantic group} $G_i^{t}$, given all seen tasks $\mathcal{T}^t=\langle \tau^1, \cdots, \tau^t \rangle$, is a subset of tasks that are within $R$ from the centroid of the group in the semantic representation space. Formally,
\begin{equation}
G_i^{t} = \{\tau \in \mathcal{T}^t~|~ \delta(G_i^t) \leq R \},
\label{eq:macro_semantic_group}
\end{equation}
where $\delta(G_i^t)$ is the average distance\,\cite{zhang1996birch} from a semantic representation $s(\tau)$ ($\tau \in G_i^{t}$) to the group centroid, calculated as 
\begin{equation}
\delta(G_i^t) = \sqrt{ \frac{1}{|{G}_i^{t}|} \sum_{\tau \in {G}_i^{t}} \left( s(\tau) - {\rm Centroid}({G}_i^{t}) \right)^2 },
\label{eq:delta}
\end{equation}
where ${\rm Centroid}({G}_i^{t})$ is the mean vector of the semantic representations of the tasks in ${G}_i^{t}$.
\end{definition}

\noindent\textbf{Task Semantic Extraction.} 
Given a task $\tau$, we obtain a warm-up prompt $\hat{{\rm P}}^{\tau}$ by the objective goal in Eq.\,\eqref{eq:spcf_objective}. Then, the \emph{semantic representation} of a given task is extracted as specified in Definition \ref{def:task_semantic_embedding}.

\begin{definition}{\sc (Task Semantic Representation)}.
Given a warm-up prompt \( \hat{\rm P}^{\tau} \) for a task \( \tau \), the \textit{task semantic representation} ${\rm s}(\tau)$ is computed as
\begin{equation}
{\rm s}(\tau) = {\rm Normalize}\left({\rm AvgPool}(\hat{\rm P}^{\tau})\right) \in \mathbb{R}^{d},
\label{eq:task_semantic_embedding}
\end{equation}
where {\rm AvgPool($\cdot$)} pools the prompt averaged over all $p$ tokens and {\rm Normalize($\cdot$)} performs $l_2$-normalization.
\label{def:task_semantic_embedding}
\end{definition}

\smallskip
\noindent\textbf{Task Assignment.} 
Given the previous semantic groups $\mathcal{G}^{t-1}$ and the current task $\tau^t$, we obtain new groups $\mathcal{G}^t$ by using an assignment function $\mathcal{A}_R$ with a threshold $R$, i.e.,  $\mathcal{G}^t=\mathcal{A}_R(\tau^t; \mathcal{G}^{t-1})$. If \emph{neither} of the previous groups $\mathcal{G}^{t-1}$ is semantically relevant to the task $\tau^t$, a new group is initialized for $\tau^t$; otherwise, it is assigned to the closest group{\,(e.g., the assignment of $\tau^{t}$ to $G_{2}^{t}$ in Figure \ref{fig:Outline}).} Formally, $\mathcal{A}_R(\tau^t; \mathcal{G}^{t-1})$ is defined as
\begin{equation}
\small
\begin{aligned}
&\begin{cases}
\mathcal{G}^{t-1}\!\cup\!\{\tau^t\} \!\!\!\!\!&\text{if}\! ~~ \delta(G_i\!\cup\!\{\tau^t\})\!>\!R \quad \!\!\! \forall G_i\!\!\in\!\mathcal{G}^{t-1} \\
\mathcal{G}^{t-1}\backslash G^{t-1}_*\!\cup\!\{ G^{t-1}_*\!\cup\!\{ \tau^t \}\}\!\!\!\!\!&\text{if}\! ~~ \delta(G_*^{t-1}\!\cup\!\{\tau^t\}) \!\leq\! R,
\end{cases}
\label{eq:mac_assign}
\end{aligned}
\end{equation}
where $G_*^{t-1}=\arg\min_{G_i^{t-1} \in \mathcal{G}^{t-1}}\delta(G_i^{t-1}\cup\{\tau^t\})$.
\smallskip

\subsubsection{Prospective Prompt Collection}

To avoid accessing historical tasks, we need to keep a repository of \emph{prospective prompts} so that a refined semantic group is equipped with the prompt that has been adapted to all tasks within the refined semantic group. Thus, this process requires finding \emph{prospective semantic groups} that likely appear in the future by semantic refinement. 
We note that the prompts for the prospective semantic groups should be learned and reserved for each incoming task because accessing previous tasks is restricted in CL.
{For example, in Figure \ref{fig:Outline}, the prompt for the semantic group $G^{t}_2$ has been adapted to the tasks $\tau^{2}$ and $\tau^{t}$; however, the prompt that has been adapted to $\tau^2$, $\tau^{3}$ and $\tau^{t}$ is also proactively reserved because these three tasks can belong to a certain prospective semantic group $\hat{G}^{t}_{1,1}$ through future refinement.}



To efficiently derive prospective semantic groups, we divide the entire set of tasks into a set of \emph{neighboring task sets} ${\mathcal{N}}^{t}=\{N^t_1, \cdots, N^t_{|{\mathcal{N}}^{t}|}\}$, where each neighboring task set contains the tasks close to one another\,(e.g., {$\tau^1$, $\tau^2$, $\tau^3$, and $\tau^{t}$ in $N_{1}^{t}$} in Figure \ref{fig:Outline}). The neighboring task sets are identified using the same assignment function $\mathcal{A}_{\gamma R}(\tau_t; \mathcal{N}^{t-1})$ with a larger threshold $\gamma R$ where $\gamma>1$.

\noindent\textbf{Prospective Semantic Groups.} 
The prospective semantic groups are the groups of similar tasks, which do \emph{not} exist in the current set $\mathcal{G}^t$ of semantic groups. Thus, in order to acquire different groupings, we apply a different clustering method such as $k$-means to each of the neighboring task sets\,\cite{bradley1998refining}. This clustering procedure is very efficient because the scope of clustering is locally confined to each neighboring task set.

Specifically, the $k$-means clustering method is executed multiple times for each neighboring task set, and the resulting clusters are added to the set $\mathcal{\hat{G}}$ of prospective semantic groups. That is,
a set of prospective semantic groups from a neighboring task set $N^t_i$ is defined by
\begin{equation}
\begin{gathered}
\{\hat{G}_{i,1}^{t},\cdots,\hat{G}_{i,|{N}_i^{t}|}^{t}\} = \text{\emph{k}-means}(N_i^t).  
\end{gathered}
\label{eq:Preparatory groups}
\end{equation}
{For example, in Figure \ref{fig:Outline}, $ \{\hat{G}^t_{1,1}, \hat{G}^t_{1,2}\}$ is derived from $N^t_1$.} While executing the $k$-means clustering method, the number $k$ of clusters is determined to maximize the silhouette score\,\cite{rousseeuw1987silhouettes}, which is a widely-used clustering quality indicator, as done by \citealt{sil_clusters}.  

Last, each group ${\hat{G}_{i,j}^{t}}$ is equipped with a \textit{prospective prompt} that is trained on participating tasks and further adapted to a new task. {Again in Figure \ref{fig:Outline}, the prompt for $\hat{G}^t_{1,1}$ is proactively adapted to $\tau^2$, $\tau^{3}$ and $\tau^{t}$} and reserved, even though it is not yet used for CL.
Please see Appendix \ref{app_sec:prop_pseudocode} for the details and complexity analysis.

\subsection{Step II: Semantic Refinement}  
\label{subsec:mic}

\algname{} evaluates whether group refinement is necessary and proceeds with the refinement when it is beneficial.

\noindent\textbf{Refinement Criterion.} Intuitively, we examine the contribution of the new task to the reduction of semantic groups. Since each semantic group has the same coverage\,(i.e., the threshold $R$), a set of fewer semantic groups implies superior generalization of the underlying semantic contexts, thereby optimizing our objective goal in Eq.~\eqref{eq:obj}. That is, the goal of refinement is to achieve $|\mathcal{G}^{t}|\leq|\mathcal{G}^{{t-1}}|$. An example of this refinement is shown in Figure \ref{fig:Outline}, {where the three semantic groups $G^{t^\prime}_1$, $G^{t^\prime}_2$, and $G^{t^\prime}_3$ are reduced to the two semantic groups $\hat{G}^t_{1,1}$ and $\hat{G}^t_{1,2}$}.

Here, it is important to mitigate the influence of task order inherent to the CL environment, since the order of data insertion influences online grouping outcomes\,\cite{zhang1996birch, bradley1998refining}. Accordingly, we implement simulation-based clustering techniques\,\cite{celebi2013comparative, gama2014survey} that permute task orders within each set $N_i^t$ during grouping to identify the one with the fewest semantic groups. Formally, the optimal task order, yielding the fewest groups, is determined by
\begin{equation}
\Bar{N}_i^{t} = \arg\min_{\Bar{N}_i^{t}}|\Bar{\mathcal{A}}_{R}(\Bar{N}_i^{t} \big)| \Hquad \textit{\text{s.t.}} \Hquad \Bar{\mathcal{A}}_{R}(\Bar{N}_i^{t}) \subseteq \hat{\mathcal{G}} ,
\vspace*{-0.1cm}
\label{eq:min_group}
\end{equation}
where $\Bar{N}_i^{t}$ is a set of the arbitrarily ordered tasks from $N_i^t$ and
$\Bar{\mathcal{A}}_{R}({N}_i^t)$ conducts grouping by sequentially applying the assignment function $\mathcal{A}_{R}$, i.e.,  $\Bar{\mathcal{A}}_{R}({N}_i^t) = \mathcal{A}_{R}(\tau^t; \mathcal{A}_{R}(\tau^{t-1}; \ldots \mathcal{A}_{R}(\tau^1; N_i^0) \ldots))$. In addition, we add the constraint to ensure that the resulting groups are matched with existing prospective semantic groups $\hat{\mathcal{G}}$ to successfully retrieve their corresponding prompts.

{For example, in Figure \ref{fig:Outline}, the tasks have been originally received in the order of $\tau^1$, $\tau^2$, $\tau^3$, $\tau^t$, and $\tau^{t^\prime}$, resulting in the three semantic groups in $N^{t^\prime}_1$ at $t^\prime$; however, by the simulation of other orders, the order of $\tau^{t^\prime}$, $\tau^3$, $\tau^t$, $\tau^2$, and $\tau^1$ leads to only two semantic groups, which triggers the refinement of these three semantic groups.
}

\noindent\textbf{Refinement.}
If permuted tasks $\Bar{N}_i^{t}$, derived from Eq.~\eqref{eq:min_group}, results in group reduction, i.e., $|\Bar{\mathcal{A}}_{R}(\Bar{N}_i^{t})|<|\Bar{\mathcal{A}}_{R}(\Bar{N}_i^{t-1})|$, then the tasks of ${N_i^{t}}$ are refined as $\Bar{\mathcal{A}}_{R}(\Bar{N}_i^{t})$; otherwise, the refinement is not performed. Following the refinement, their prompts are retrieved from the prospective semantic groups $\hat{\mathcal{G}}$ by further adapting to the current task $\tau^t$.

\subsection{Prompt Tuning and Inference}  
\label{subsec:tr_inf}

\smallskip
\noindent\textbf{Semantic Group-based Training.}
\algname{} proactively trains the prompts and keys for semantic groups in prospective semantic groups $\hat{\mathcal{G}}$ including a new task $\tau^t$, to prepare forthcoming refinements.
Let $\theta^t$ be a set of prompt-key pairs for the groups containing the task $\tau^t$,
\begin{equation}
    \theta^{t} = \{ ({\rm P}_{\hat{G}_{i,j}^{t}}, {\rm k}_{\hat{G}_{i,j}^{t}}) \mid \tau^t \in {\hat{G}_{i,j}^t} \in \hat{\mathcal{G}} \}.
\label{eq:prompts4tuning}
\end{equation}
Then, the training objective is to optimize $\theta^t$ by
\begin{equation}
\small
\min_{\theta^{t}, \phi} \mathbb{E}_{({\rm P}, {\rm k})\sim \text{U}(\theta^t)}\!\left[ \ell_{ce} (f_\phi(f([x^{t}; {\rm P}])),\! y^{t})\! + \!\lambda{\rm d}(f(x^{t}),\!{\rm k}) \right]\!, 
\label{eq:our_objective}
\end{equation}
where $({\rm P}, {\rm k})$ is sampled uniformly from the set $\theta^{t}$.


\noindent\textbf{Semantic Group-based Inference.} 
Given a test instance $x$, we find the matching prompt ${\rm P}$ by Eq.~\eqref{eq:preliminary-inference} but from the set $\mathcal{G}^{t}$ of semantic groups. Specifically, the prompt of ${i^*}$-th semantic group is chosen as the best for $x$ if its key is the most similar to $f(x)$,
\begin{equation}
{i^*} = \arg\min_{i} {\rm d}(f(x), {\rm k}_{G_{i}^t}).
\label{eq:inference}
\end{equation}
\section{Evaluation}
\label{sec:evaluation}



\subsection{Experiment Setting}
\noindent\textbf{Dataset Preparation.}
Our evaluation of \algname{} focuses on its efficacy across a spectrum of task semantic shifts, \emph{uniform} and \emph{varying} shifts. CL scenarios are categorized into uniformly mild, uniformly abrupt, and varying shifts. Visual comparisons between uniformly mild and abrupt scenarios are provided in Appendix \ref{app_sec:vtab_dataset}.

\def\arraystretch{1.4}
\begin{table*}[t!]
\small
\centering
\caption{Performance\,(last accuracy) comparison of \algname{} against CL baselines using prompt tuning across uniform and varying shift scenarios. ``{\it Avg.~Improv.}'' denotes the relative improvement of \algname{} over the average of the five baselines, and ``{\it \#~Semantics}'' the number of semantic groups by \algname{}.  
The best and second-best values are highlighted in bold and underlined, respectively.
Note that \algname{} is even better than or comparable to the baselines for the uniform shifts which are \emph{not} our sweet spots.
}
\vspace*{0.3cm}
\resizebox{0.9\linewidth}{!}{%
\begin{tabular}[c]
{@{}cc|ccccc|cccc@{}}
\toprule
\multirow{2}{*}{\hspace{-0.0cm}\makecell[c]{Shift\\Scenarios}\hspace{-0.0cm}}&\multirow{2}{*}{\makecell[c]{CL\\Datasets}} 
& \multicolumn{5}{c|}{{ {Prompt Tuning CL Algorithms}}} 
& \multicolumn{3}{c}{{ {\algname{} }}}\\
& & \multicolumn{1}{c}{\hspace{-0.18cm}{L2P}\hspace{-0.18cm}}
& \multicolumn{1}{c}{\hspace{-0.18cm}{VPT}\hspace{-0.18cm}}
& \multicolumn{1}{c}{\hspace{-0.18cm}{LAE}\hspace{-0.18cm}}
& \multicolumn{1}{c}{\hspace{-0.18cm}{DP}\hspace{-0.18cm}}
& \multicolumn{1}{c|}{\hspace{-0.18cm}{S-Prompts}\hspace{-0.18cm}}
& \multicolumn{1}{c}{\hspace{-0.05cm}{\textbf{\algname{}}}\hspace{-0.18cm}}
& \multicolumn{1}{c}{\hspace{-0.18cm}{\textit{Avg. Improv.}}\hspace{-0.18cm}}
& \multicolumn{1}{c}{\hspace{-0.18cm}{\textit{\#~Semantics}}\hspace{-0.18cm}}
\\ \midrule

\multirow{6}{*}{\hspace{-0.0cm}{\textbf{Varying}}\hspace{-0.18cm}} 
& \multirow{1}{*}{\makecell[c]{VTAB-Sim25}}
& \begin{tabular}[c]{@{}c@{}}{ \hspace{-0.18cm}{35.77}\scalebox{0.99}\,($\pm$0.40)\hspace{-0.18cm} }\end{tabular}
& \begin{tabular}[c]{@{}c@{}}{ \hspace{-0.18cm}{36.38}\scalebox{0.99}\,($\pm$0.30)\hspace{-0.18cm} }\end{tabular}
& \begin{tabular}[c]{@{}c@{}}{ \hspace{-0.18cm}{35.29}\scalebox{0.99}\,($\pm$0.42)\hspace{-0.18cm} }\end{tabular}
& \begin{tabular}[c]{@{}c@{}}{ \hspace{-0.18cm}{35.77}\scalebox{0.99}\,($\pm$0.34)\hspace{-0.18cm} }\end{tabular}
& \begin{tabular}[c]{@{}c@{}}{ \hspace{-0.18cm}{\underline{36.61}}\scalebox{0.99}\,($\pm$0.42)\hspace{-0.18cm} }\end{tabular}
& \begin{tabular}[c]{@{}c@{}}{ \hspace{-0.18cm}{\textbf{38.29}}\scalebox{0.99}\,($\pm$0.36)\hspace{-0.18cm} }\end{tabular}
& \begin{tabular}[c]{@{}c@{}}{ \hspace{-0.18cm}{\textit{6.49}}\%\hspace{-0.18cm} }\end{tabular}
& \multirow{1}{*}{\makecell[c]{16.0}}
\\ 
& \multirow{1}{*}{\makecell[c]{VTAB-Sim50}}
& \begin{tabular}[c]{@{}c@{}}{ \hspace{-0.18cm}{36.06}\scalebox{0.99}\,($\pm$0.39)\hspace{-0.18cm} }\end{tabular}
& \begin{tabular}[c]{@{}c@{}}{ \hspace{-0.18cm}{36.15}\scalebox{0.99}\,($\pm$0.32)\hspace{-0.18cm} }\end{tabular}
& \begin{tabular}[c]{@{}c@{}}{ \hspace{-0.18cm}{35.23}\scalebox{0.99}\,($\pm$0.61)\hspace{-0.18cm} }\end{tabular}
& \begin{tabular}[c]{@{}c@{}}{ \hspace{-0.18cm}{35.57}\scalebox{0.99}\,($\pm$0.40)\hspace{-0.18cm} }\end{tabular}
& \begin{tabular}[c]{@{}c@{}}{ \hspace{-0.18cm}{\underline{36.25}}\scalebox{0.99}\,($\pm$0.45)\hspace{-0.18cm} }\end{tabular}
& \begin{tabular}[c]{@{}c@{}}{ \hspace{-0.18cm}{\textbf{37.92}}\scalebox{0.99}\,($\pm$0.47)\hspace{-0.18cm} }\end{tabular}
& \begin{tabular}[c]{@{}c@{}}{ \hspace{-0.18cm}{\textit{5.78}}\%\hspace{-0.18cm} }\end{tabular}
& \multirow{1}{*}{\makecell[c]{13.4}}
\\ 
& \multirow{1}{*}{\makecell[c]{VTAB-Sim75}}
& \begin{tabular}[c]{@{}c@{}}{ \hspace{-0.18cm}{35.66}\scalebox{0.99}\,($\pm$0.44)\hspace{-0.18cm} }\end{tabular}
& \begin{tabular}[c]{@{}c@{}}{ \hspace{-0.18cm}{36.06}\scalebox{0.99}\,($\pm$0.42)\hspace{-0.18cm} }\end{tabular}
& \begin{tabular}[c]{@{}c@{}}{ \hspace{-0.18cm}{35.66}\scalebox{0.99}\,($\pm$0.60)\hspace{-0.18cm} }\end{tabular}
& \begin{tabular}[c]{@{}c@{}}{ \hspace{-0.18cm}{\underline{36.35}}\scalebox{0.99}\,($\pm$0.46)\hspace{-0.18cm} }\end{tabular}
& \begin{tabular}[c]{@{}c@{}}{ \hspace{-0.18cm}{35.76}\scalebox{0.99}\,($\pm$0.53)\hspace{-0.18cm} }\end{tabular}
& \begin{tabular}[c]{@{}c@{}}{ \hspace{-0.18cm}{\textbf{37.59}}\scalebox{0.99}\,($\pm$0.37)\hspace{-0.18cm} }\end{tabular}
& \begin{tabular}[c]{@{}c@{}}{ \hspace{-0.18cm}{\textit{4.72}}\%\hspace{-0.18cm} }\end{tabular}
& \multirow{1}{*}{\makecell[c]{12.8}}
\\ \addlinespace[0.3ex]\cline{2-11}\addlinespace[0.5ex]
& \multirow{1}{*}{\makecell[c]{VTAB-Rec2}}
& \begin{tabular}[c]{@{}c@{}}{ \hspace{-0.18cm}{52.18}\scalebox{0.99}\,($\pm$0.13)\hspace{-0.18cm} }\end{tabular}
& \begin{tabular}[c]{@{}c@{}}{ \hspace{-0.18cm}{51.84}\scalebox{0.99}\,($\pm$0.12)\hspace{-0.18cm} }\end{tabular}
& \begin{tabular}[c]{@{}c@{}}{ \hspace{-0.18cm}{52.34}\scalebox{0.99}\,($\pm$0.37)\hspace{-0.18cm} }\end{tabular}
& \begin{tabular}[c]{@{}c@{}}{ \hspace{-0.18cm}{51.66}\scalebox{0.99}\,($\pm$0.15)\hspace{-0.18cm} }\end{tabular}
& \begin{tabular}[c]{@{}c@{}}{ \hspace{-0.18cm}{\underline{53.28}}\scalebox{0.99}\,($\pm$0.16)\hspace{-0.18cm} }\end{tabular}
& \begin{tabular}[c]{@{}c@{}}{ \hspace{-0.18cm}{\textbf{56.40}}\scalebox{0.99}\,($\pm$0.25)\hspace{-0.18cm} }\end{tabular}
& \begin{tabular}[c]{@{}c@{}}{ \hspace{-0.18cm}{\textit{7.93}}\%\hspace{-0.18cm} }\end{tabular}
& \multirow{1}{*}{\makecell[c]{5.0}}
\\ 
& \multirow{1}{*}{\makecell[c]{VTAB-Rec5}}
& \begin{tabular}[c]{@{}c@{}}{ \hspace{-0.18cm}{52.68}\scalebox{0.99}\,($\pm$0.24)\hspace{-0.18cm} }\end{tabular}
& \begin{tabular}[c]{@{}c@{}}{ \hspace{-0.18cm}{52.76}\scalebox{0.99}\,($\pm$0.11)\hspace{-0.18cm} }\end{tabular}
& \begin{tabular}[c]{@{}c@{}}{ \hspace{-0.18cm}{\underline{54.06}}\scalebox{0.99}\,($\pm$0.26)\hspace{-0.18cm} }\end{tabular}
& \begin{tabular}[c]{@{}c@{}}{ \hspace{-0.18cm}{51.66}\scalebox{0.99}\,($\pm$0.16)\hspace{-0.18cm} }\end{tabular}
& \begin{tabular}[c]{@{}c@{}}{ \hspace{-0.18cm}{52.82}\scalebox{0.99}\,($\pm$0.36)\hspace{-0.18cm} }\end{tabular}
& \begin{tabular}[c]{@{}c@{}}{ \hspace{-0.18cm}{\textbf{56.86}}\scalebox{0.99}\,($\pm$0.33)\hspace{-0.18cm} }\end{tabular}
& \begin{tabular}[c]{@{}c@{}}{ \hspace{-0.18cm}{\textit{7.72}}\%\hspace{-0.18cm} }\end{tabular}
& \multirow{1}{*}{\makecell[c]{5.6}}
\\ 
& \multirow{1}{*}{\makecell[c]{VTAB-Rec10}}
& \begin{tabular}[c]{@{}c@{}}{ \hspace{-0.18cm}{52.20}\scalebox{0.99}\,($\pm$0.81)\hspace{-0.18cm} }\end{tabular}
& \begin{tabular}[c]{@{}c@{}}{ \hspace{-0.18cm}{51.70}\scalebox{0.99}\,($\pm$0.44)\hspace{-0.18cm} }\end{tabular}
& \begin{tabular}[c]{@{}c@{}}{ \hspace{-0.18cm}{\underline{52.34}}\scalebox{0.99}\,($\pm$0.35)\hspace{-0.18cm} }\end{tabular}
& \begin{tabular}[c]{@{}c@{}}{ \hspace{-0.18cm}{50.80}\scalebox{0.99}\,($\pm$0.58)\hspace{-0.18cm} }\end{tabular}
& \begin{tabular}[c]{@{}c@{}}{ \hspace{-0.18cm}{47.72}\scalebox{0.99}\,($\pm$0.20)\hspace{-0.08cm} }\end{tabular}
& \begin{tabular}[c]{@{}c@{}}{ \hspace{-0.18cm}{\textbf{54.22}}\scalebox{0.99}\,($\pm$0.52)\hspace{-0.18cm} }\end{tabular}
& \begin{tabular}[c]{@{}c@{}}{ \hspace{-0.18cm}{\textit{6.54}}\%\hspace{-0.18cm} }\end{tabular}
& \multirow{1}{*}{\makecell[c]{6.2}}
\\ \midrule
\multirow{2}{*}{\hspace{-0.0cm}{\parbox{1.5cm}{\centering\textbf{Uniformly Mild}}}\hspace{-0.18cm}} 
& \multirow{1}{*}{\makecell[c]{ImageNet-R}}
& \begin{tabular}[c]{@{}c@{}}{ \hspace{-0.18cm}{68.05}\scalebox{0.99}\,($\pm$0.11)\hspace{-0.18cm} }\end{tabular}
& \begin{tabular}[c]{@{}c@{}}{ \hspace{-0.18cm}{69.31}\scalebox{0.99}\,($\pm$0.09)\hspace{-0.18cm} }\end{tabular}
& \begin{tabular}[c]{@{}c@{}}{ \hspace{-0.18cm}{\textbf{70.11}}\scalebox{0.99}\,($\pm$0.12)\hspace{-0.18cm} }\end{tabular}
& \begin{tabular}[c]{@{}c@{}}{ \hspace{-0.18cm}{67.81}\scalebox{0.99}\,($\pm$0.19)\hspace{-0.18cm} }\end{tabular}
& \begin{tabular}[c]{@{}c@{}}{ \hspace{-0.18cm}{65.89}\scalebox{0.99}\,($\pm$0.10)\hspace{-0.08cm} }\end{tabular}
& \begin{tabular}[c]{@{}c@{}}{ \hspace{-0.18cm}{\underline{69.45}}\scalebox{0.99}\,({$\pm$0.14})\hspace{-0.18cm} }\end{tabular}
& \begin{tabular}[c]{@{}c@{}}{ \hspace{-0.18cm}{\textit{1.83}}\%\hspace{-0.18cm} }\end{tabular}
& \multirow{1}{*}{\makecell[c]{1.0}}
\\ 
& \multirow{1}{*}{\makecell[c]{CIFAR100}}
& \begin{tabular}[c]{@{}c@{}}{ \hspace{-0.18cm}{84.55}\scalebox{0.99}\,($\pm$0.08)\hspace{-0.18cm} }\end{tabular}
& \begin{tabular}[c]{@{}c@{}}{ \hspace{-0.18cm}{\textbf{85.34}}\scalebox{0.99}\,($\pm$0.15)\hspace{-0.18cm} }\end{tabular}
& \begin{tabular}[c]{@{}c@{}}{ \hspace{-0.18cm}{85.16}\scalebox{0.99}\,($\pm$0.11)\hspace{-0.18cm} }\end{tabular}
& \begin{tabular}[c]{@{}c@{}}{ \hspace{-0.18cm}{84.79}\scalebox{0.99}\,($\pm$0.14)\hspace{-0.18cm} }\end{tabular}
& \begin{tabular}[c]{@{}c@{}}{ \hspace{-0.18cm}{83.73}\scalebox{0.99}\,($\pm$0.10)\hspace{-0.08cm} }\end{tabular}
& \begin{tabular}[c]{@{}c@{}}{ \hspace{-0.18cm}{\underline{85.31}}\scalebox{0.99}\,($\pm$0.11)\hspace{-0.18cm} }\end{tabular}
& \begin{tabular}[c]{@{}c@{}}{ \hspace{-0.18cm}{\textit{0.71}}\%\hspace{-0.18cm} }\end{tabular}
& \multirow{1}{*}{\makecell[c]{1.0}}
\\
\midrule
\multirow{2}{*}{\hspace{-0.0cm}{\parbox{1.5cm}{\centering\textbf{Uniformly Abrupt}}}\hspace{-0.18cm}} 
& \multirow{1}{*}{\makecell[c]{VTAB-19T}}
& \begin{tabular}[c]{@{}c@{}}{ \hspace{-0.18cm}{28.23}\scalebox{0.99}\,($\pm$0.14)\hspace{-0.18cm} }\end{tabular}
& \begin{tabular}[c]{@{}c@{}}{ \hspace{-0.18cm}{28.11}\scalebox{0.99}\,($\pm$0.20)\hspace{-0.18cm} }\end{tabular}
& \begin{tabular}[c]{@{}c@{}}{ \hspace{-0.18cm}{26.71}\scalebox{0.99}\,($\pm$0.29)\hspace{-0.18cm} }\end{tabular}
& \begin{tabular}[c]{@{}c@{}}{ \hspace{-0.18cm}{28.48}\scalebox{0.99}\,($\pm$0.07)\hspace{-0.18cm} }\end{tabular}
& \begin{tabular}[c]{@{}c@{}}{ \hspace{-0.18cm}{\underline{30.83}}\scalebox{0.99}\,($\pm$0.08)\hspace{-0.08cm} }\end{tabular}
& \begin{tabular}[c]{@{}c@{}}{ \hspace{-0.18cm}{\textbf{32.39}}\scalebox{0.99}\,($\pm$0.32)\hspace{-0.18cm} }\end{tabular}
& \begin{tabular}[c]{@{}c@{}}{ \hspace{-0.18cm}{\textit{14.00}}\%\hspace{-0.18cm} }\end{tabular}
& \multirow{1}{*}{\makecell[c]{18.0}}
\\ 
& \multirow{1}{*}{\makecell[c]{VTAB-5T}}
& \begin{tabular}[c]{@{}c@{}}{ \hspace{-0.18cm}{34.31}\scalebox{0.99}\,($\pm$0.04)\hspace{-0.18cm} }\end{tabular}
& \begin{tabular}[c]{@{}c@{}}{ \hspace{-0.18cm}{34.03}\scalebox{0.99}\,($\pm$0.05)\hspace{-0.18cm} }\end{tabular}
& \begin{tabular}[c]{@{}c@{}}{ \hspace{-0.18cm}{35.30}\scalebox{0.99}\,($\pm$0.41)\hspace{-0.18cm} }\end{tabular}
& \begin{tabular}[c]{@{}c@{}}{ \hspace{-0.18cm}{34.17}\scalebox{0.99}\,($\pm$0.04)\hspace{-0.18cm} }\end{tabular}
& \begin{tabular}[c]{@{}c@{}}{ \hspace{-0.18cm}{\underline{38.27}}\scalebox{0.99}\,($\pm$0.18)\hspace{-0.08cm} }\end{tabular}
& \begin{tabular}[c]{@{}c@{}}{ \hspace{-0.18cm}{\textbf{38.67}}\scalebox{0.99}\,($\pm$0.15)\hspace{-0.18cm} }\end{tabular}
& \begin{tabular}[c]{@{}c@{}}{ \hspace{-0.18cm}{\textit{10.02}}\%\hspace{-0.18cm} }\end{tabular}
& \multirow{1}{*}{\makecell[c]{5.0}}
\\
\bottomrule
\end{tabular}
}
\label{tbl:overall_perf}
\vspace{-0.0cm}
\end{table*}

\noindent{\underline{Uniformly Mild Scenario:}} We utilize the \emph{CIFAR-100}\,\cite{krizhevsky2009learning} and \emph{ImageNet-R}\,\cite{hendrycks2021many} datasets to simulate uniformly mild shifts between tasks.  
By randomly subdividing these datasets into task subsets, we maintain a high degree of similarity across tasks. Following \citealt{wang2022learning, wang2022dualprompt}, we split {CIFAR-100} and {ImageNet-R} into 10 tasks, with disjoint class sets resulting in 10 and 20 classes per task, respectively.

\noindent{\underline{Uniformly Abrupt Scenario:}} We employ the Visual Task Adaptation Benchmark\,(VTAB)\,\cite{zhai2019large} with its 19 distinct datasets to represent uniformly abrupt task shifts. In this benchmark referred to as \emph{VTAB-19T}, each dataset within VTAB is treated as a separate task. 
Additionally, to simulate severe abrupt shifts, we construct \textit{VTAB-5T}, which comprises the five most semantically distinct datasets.

\noindent{\underline{Varying Scenario:}} We construct two CL benchmarks, \textit{VTAB-SimS} and \textit{VTAB-RecR}, based on VTAB-19T. They are designed to contain the shifts between tasks that are either semantically similar or dissimilar. For the selection of semantically dissimilar tasks, we draw them from the existing VTAB tasks. For the generation of similar tasks, we employ task merging\,\cite{koh2022online, bang2021rainbow} for VTAB-SimS and task recurrence\,\cite{cossu2022class} for VTAB-RecR. Specifically, in VTAB-SimS, tasks overlap in `S'\% of their data instances, whereas, in VTAB-RecR, tasks are repeated `R' times. We repeat five tasks for scalable evaluation, resulting in five unique semantics within the VTAB-RecR benchmark. Three levels of similarity are tested using the parameters S:\{25, 50, 75\} and R:\{2, 5, 10\}, and further details can be found in Appendix \ref{app_sec:vtab_dataset}.

\smallskip
\noindent\textbf{Algorithms and Metric.}
\algname{} is compared against representative rehearsal-free CL methods: L2P\,\cite{wang2022learning}, VPT\,\cite{jia2022visual}, DP\,\cite{wang2022dualprompt}, S-Prompts\,\cite{wang2022sprompts}, and LAE\,\cite{gao2023unified}. For evaluation, we adopt a widely-used performance metric, 
the \emph{last accuracy} $A_{last} = \frac{1}{T}\Sigma_{i=1}^{T}A_{T, i}$, where $A_{T, i}$ is the accuracy of the model on $i$-th task after learning the $T$-th task sequentially. 
Given that the last accuracy encompasses both learning adaptability and memory retention, it serves as a comprehensive measure of CL performance\,\cite{smith2023coda}. For reliability, we run every experiment \textit{five} times with different random seeds and report the average value with the standard error. 

\smallskip
\noindent\textbf{Implementation Details.}
All baseline methods are implemented using the publicly accessible CL framework\,\footnote{https://github.com/JH-LEE-KR/dualprompt-pytorch}, following \cite{gao2023unified}.
A ViT-B/16\,\cite{dosovitskiy2021an} pre-trained on ImageNet-1k is utilized as the common backbone across all methods, with prompts appended to the initial five transformer layers. 
Optimization is carried out using the Adam optimizer, setting a batch size of 128 and a learning rate of 0.025. For LAE, as per the original paper, a reduced learning rate of 0.005 is employed. The epoch counts are fixed to 5 and 50 for the universal and specific methods, respectively, reflecting their respective convergence patterns.
For \algname{}, the configuration includes 150 warm-up iterations, with the assignment threshold $R = 0.4$ and $\gamma = 3/2$ to maintain consistent relative scaling\,\cite{kozawa2015parallel} across all datasets.
All methods are implemented using PyTorch 1.12.1 and Timm 0.8.0 and tested on two NVIDIA RTX 3080 GPUs, and the source code is publicly available at {\url{https://github.com/kaist-dmlab/AdaPromptCL}}.

\smallskip
\subsection{Main Results}

\noindent\textbf{Varying Scenarios.}
Table \ref{tbl:overall_perf} presents a comparative analysis of \algname{} against universal and specific prompting methods. 
In the varying shift scenarios, which we emphasize on, \algname{} outperforms the other baseline methods across all datasets. The empirical results indicate that \algname{} attains an average improvement of 6.35\% and 6.91\% compared to LAE and S-Prompts across varying scenarios, respectively. Unlike \algname{}, the baseline methods fail to adjust to semantic shifts, resulting in the use of prompts that are either inadequate or excessive, which consequently diminishes their efficacy.

\smallskip
\noindent\textbf{Uniformly Mild and Abrupt Scenarios.}
In general, \algname{} demonstrates consistently high performance in the uniformly mild and abrupt shift scenarios. Specifically, \algname{} matches the performance of baselines with universal prompts in the uniformly mild scenario and those with specific prompts in the uniformly abrupt scenario, where each is considered the optimal approach. In contrast, the efficacy of the other baselines varies considerably, according to the extent of semantic shifts.


Interestingly, we observe that \algname{} generates the averages of 1 and 18 prompts, respectively, in the uniformly mild scenario\,(ImageNet-R) and the uniformly abrupt scenario\,(VTAB-19T). These values align with the optimal number of prompts employed in each scenario when universal and specific prompting methods are individually applied. This result demonstrates \algname{}'s capability to flexibly adapt its prompting architecture to the degree of task semantic shifts, leveraging the benefits of both universal and specific prompts within a single model. Please see Appendix \ref{app_sec:additional_exp} for the results about the \emph{forgetting} metric.

\smallskip
\subsection{In-depth Analysis of Semantic Refinement}
\label{subsec:analysis_of_semantic_refinement}

\def\arraystretch{1.1}
\begin{table}[t]
\small
\centering
\vspace*{-0.2cm}
\caption{
Ablation analysis highlighting the impact of semantic refinement and proactive prompt tuning on the last accuracy. The highest values are emphasized in bold.}
\vspace*{0.3cm}
\resizebox{1\linewidth}{!}{%
\begin{tabular}[c]
{@{}l||c|c|cc@{}}
\toprule
\multicolumn{1}{c||}{{Shift Scenarios}} 
& \multicolumn{1}{c|}{\textbf{No Refine}}
& \multicolumn{1}{c|}{\textbf{Avg Merge}}
& \multicolumn{1}{c}{\textbf{\algname{}}}

\\ \toprule
\multirow{1}{*}{\textbf{Uniformly Mild}\,\scalebox{0.80}{(ImageNet-R)}}                                
& \begin{tabular}[c]{@{}c@{}}{ \hspace{-0.1cm}{85.31}\scalebox{0.99}\,($\pm$0.11)\hspace{-0.1cm} }\end{tabular}
& \begin{tabular}[c]{@{}c@{}}{ \hspace{-0.1cm}{85.31}\scalebox{0.99}\,($\pm$0.11)\hspace{-0.1cm} }\end{tabular}
& \begin{tabular}[c]{@{}c@{}}{ \hspace{-0.1cm}{85.31}\scalebox{0.99}\,($\pm$0.11)\hspace{-0.1cm} }\end{tabular}
\\

\multirow{1}{*}{\textbf{Uniformly Abrupt}\,\scalebox{0.80}{(VTAB-19T)}}                                                    
& \begin{tabular}[c]{@{}c@{}}{ \hspace{-0.1cm}{31.25}\scalebox{0.99}\,($\pm$0.20)\hspace{-0.1cm} }\end{tabular}
& \begin{tabular}[c]{@{}c@{}}{ \hspace{-0.1cm}{30.08}\scalebox{0.99}\,($\pm$0.26)\hspace{-0.1cm} }\end{tabular}
& \begin{tabular}[c]{@{}c@{}}{ \hspace{-0.1cm}{\textbf{32.39}}\scalebox{0.99}\,($\pm$0.32)\hspace{-0.1cm} }\end{tabular}
\\
\multirow{1}{*}{\textbf{Varying}\,\scalebox{0.80}{(VTAB-Rec10)}}                                     
& \begin{tabular}[c]{@{}c@{}}{ \hspace{-0.1cm}{51.80}\scalebox{0.99}\,($\pm$0.62)\hspace{-0.1cm} }\end{tabular}
& \begin{tabular}[c]{@{}c@{}}{ \hspace{-0.1cm}{52.92}\scalebox{0.99}\,($\pm$1.06)\hspace{-0.1cm} }\end{tabular}
& \begin{tabular}[c]{@{}c@{}}{ \hspace{-0.1cm}{\textbf{54.22}}\scalebox{0.99}\,($\pm$0.52)\hspace{-0.1cm} }\end{tabular}
\\
\midrule
\multirow{1}{*}{{\textit{Average Degradation.}}\hspace{-0.15cm}}            
& \begin{tabular}[c]{@{}c@{}}{{2.8\%}\hspace{-1cm}}\end{tabular}
& \begin{tabular}[c]{@{}c@{}}{{3.4\%}}\end{tabular}
& \begin{tabular}[c]{@{}c@{}}{{-}}\end{tabular}
\\
\bottomrule
\end{tabular} }
\label{tbl:ablation}
\vspace*{-0.1cm}
\end{table}
\noindent\textbf{Ablation Study.}
Our analysis focuses on the effectiveness of semantic refinement. Table \ref{tbl:ablation} compares \algname{} with its two variants across three semantic shift scenarios.
\textit{No Refine} omits the refinement process, which solely relies on the online assignment in Eq.~\eqref{eq:mac_assign} with the threshold $R$. \textit{Avg Merge} creates the prompts for the refined semantic groups using simple prompt averages, rather than adapting them to the tasks within these groups; that is, it does not perform proactive tuning of the prompts for prospective semantic groups in Eq.~\eqref{eq:Preparatory groups}.

Overall, the performance of both variants degrades when compared to \algname{}, with average decreases in accuracy of 3.55\% and 5.63\% on uniformly abrupt shifts and varying shifts, respectively.
Notably, \textit{No Refine} exhibits a greater performance drop in the varying shift scenario, highlighting its susceptibility to inaccurate groupings due to the lack of the refining mechanism.
In addition, \textit{Avg Merge} shows vulnerability in the uniformly abrupt scenario, indicating that merely averaging the parameters of prompts is inadequate to blend knowledge acquired individually on different tasks without adaptation to prior tasks. 

\noindent\textbf{Correctness of Group Refinement.}
Table \ref{tbl:Eval_semantics} presents the correctness of semantic grouping in the varying shift scenario using widely-used clustering quality metrics, adjusted rand index\,\cite{steinley2004properties} and normalized mutual information\,\cite{hubert1985comparing}. The true task-to-group assignments are known in the process of generating VTAB-RecR. Details on the metrics are available in Appendix \ref{app_sec:clst_metrics}.
Notably, \algname{} has successfully refined the groupings, reducing semantic groups from an average of 9.6 to 6.2. Moreover, the improvements in the metrics, approaching the optimal value of 1.0, further affirm its capability to enhance the precision of semantic grouping.

\def\arraystretch{1.1}
\begin{table}[t]
\small
\centering
\caption{Comparison of grouping quality between \algname{} and \textit{No Refine} variant, with superior values highlighted in bold.}
\vspace*{0.3cm}
\resizebox{1\linewidth}{!}{%
\begin{tabular}[c]
{@{}l||l|l|cc@{}}
\toprule
\multicolumn{1}{c||}{{}} 
& \multicolumn{1}{c|}{\textbf{No Refine}}
& \multicolumn{1}{c|}{\textbf{\algname{}}}
& \multicolumn{1}{c}{\textit{Reference}}

\\ \toprule
\multirow{1}{*}{\textit{\# Semantic Groups}}                                
& \begin{tabular}[c]{@{}c@{}}{ \hspace{-0.01cm}{9.60}\scalebox{0.99}\,($\pm$0.540)\hspace{-0.01cm} }\end{tabular}
& \begin{tabular}[c]{@{}c@{}}{ \hspace{-0.01cm}{\textbf{6.20}}\scalebox{0.99}\,($\pm$0.090)\hspace{-0.01cm} }\end{tabular}
& \begin{tabular}[c]{@{}c@{}}{ \hspace{-0.01cm}{5}\hspace{-0.01cm} }\end{tabular}

\\
\midrule

\multirow{1}{*}{\textit{Adj. Rand Index}}                                   
& \begin{tabular}[c]{@{}c@{}}{ \hspace{-0.01cm}{0.89}\scalebox{0.99}\,($\pm$0.031)\hspace{-0.01cm} }\end{tabular}
& \begin{tabular}[c]{@{}c@{}}{ \hspace{-0.01cm}{\textbf{0.97}}\scalebox{0.99}\,($\pm$0.005)\hspace{-0.01cm} }\end{tabular}
& \begin{tabular}[c]{@{}c@{}}{ \hspace{-0.01cm}{1}\hspace{-0.01cm} }\end{tabular}
\\
\multirow{1}{*}{\textit{Norm. Mutual Information}}                                   
& \begin{tabular}[c]{@{}c@{}}{ \hspace{-0.01cm}{0.90}\scalebox{0.99}\,($\pm$0.027)\hspace{-0.01cm} }\end{tabular}
& \begin{tabular}[c]{@{}c@{}}{ \hspace{-0.01cm}{\textbf{0.98}}\scalebox{0.99}\,($\pm$0.004)\hspace{-0.01cm} }\end{tabular}
& \begin{tabular}[c]{@{}c@{}}{ \hspace{-0.01cm}{1}\hspace{-0.01cm} }\end{tabular}
\\
\bottomrule
\end{tabular} }
\label{tbl:Eval_semantics}
\end{table}
\begin{figure}[t!]
\centering 
\includegraphics[width=0.99\linewidth]{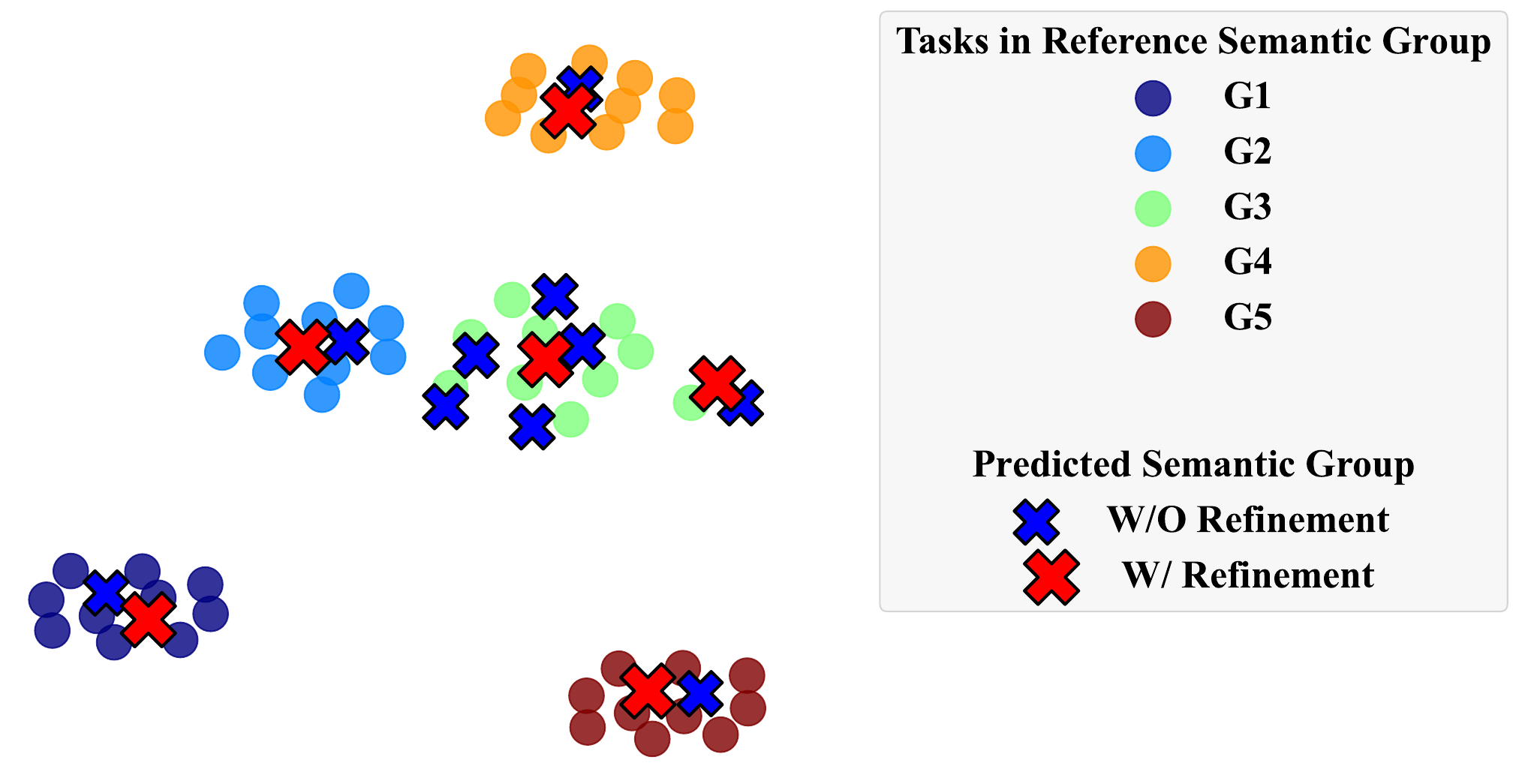}
\vspace*{-0.0cm}
\caption{t-SNE visualization on the semantic groups formed by \algname{} across 50 tasks on VTAB-Rec10. The tasks from the same semantic origin are in the same color, denoted by circles \scalebox{0.99}{\color{gray}\CIRCLE}. The symbols \textcolor{red}{\textbf{\texttimes{}}} and \textcolor{blue}{\textbf{\texttimes{}}} respectively represent the centroids of semantic groups with and without refinement.}
\vspace*{-0.0cm}
\label{fig:vis_groups}
\end{figure}

\noindent\textbf{Visualization of Semantic Grouping.}
The t-SNE visualization in Figure \ref{fig:vis_groups} offers a qualitative insight into the efficacy of the refinement process. Tasks are plotted using their semantic representations ${\rm s}(\tau)$ from Eq.~\eqref{eq:task_semantic_embedding}, and semantic groups are indicated by the centroids of the tasks within each group.
When the semantic groups are undesirably specified by assigning similar tasks to different semantic groups\,(see \textcolor{blue}{\textbf{\texttimes{}}} symbols around the green\,(G3) tasks), \algname{} successfully reduces the overgeneration of semantic groups\,(see \textcolor{red}{\textbf{\texttimes{}}} symbols). Please see Appendix \ref{app_sec:add_vis} for more scenarios.
\label{subsec:anal_micro_refine}

\subsection{Parameter Sensitivity Analysis}
\begin{figure}[t!]
\hspace*{-0.1cm} 
\includegraphics[width=1.00\columnwidth]{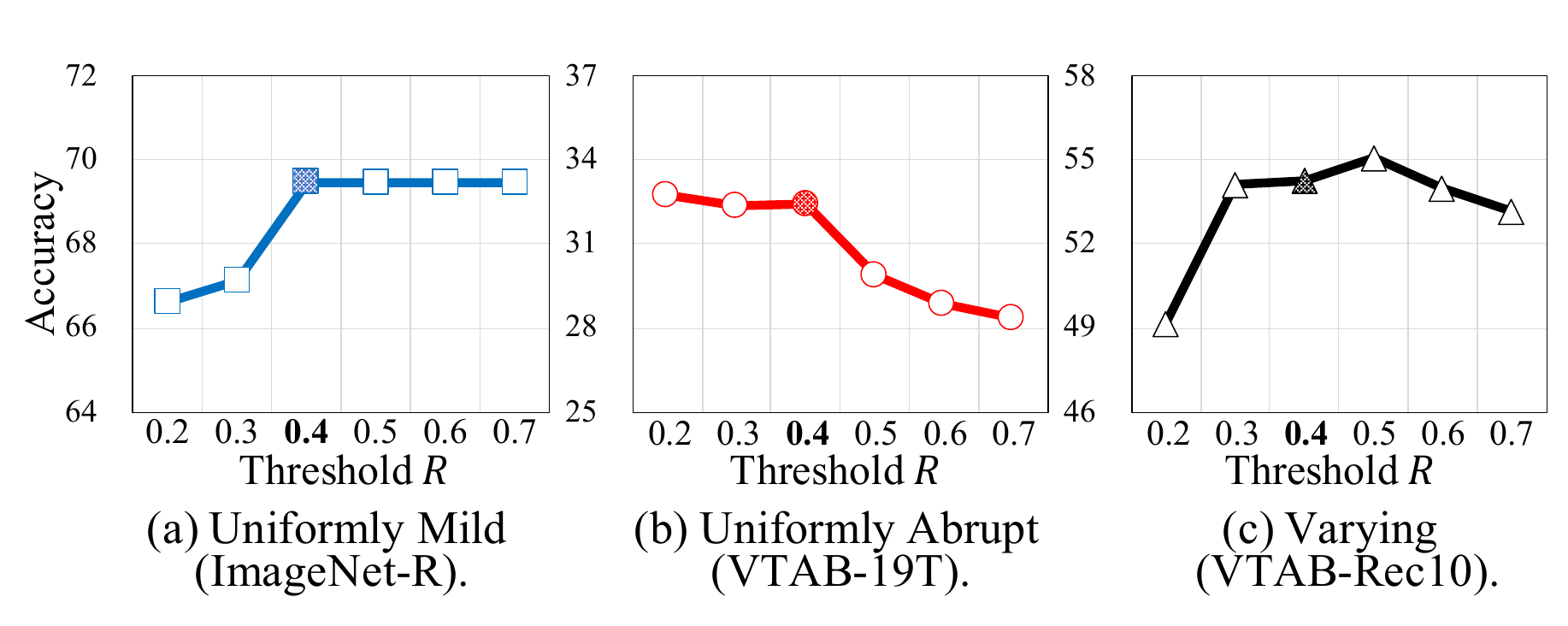}
\vspace*{-0.79cm}
\caption{Sensitivity analysis on the assignment threshold $R$.
}
\vspace*{0.0cm}
\label{fig:ablation_thrh}
\end{figure}

Figure \ref{fig:ablation_thrh} presents sensitivity analysis on \algname{}'s main hyperparameter, the semantic assignment threshold\,($R$) in Eq.~\eqref{eq:mac_assign}. 
Its impact on the last accuracy across the three scenarios---uniformly mild, uniformly abrupt, and varying---is analyzed with fixing  $\gamma$ (for neighboring task sets) to $3/2$. A lower $R$ results in smaller semantic groups, akin to a specific prompting, whereas a higher $R$ leads to larger semantic groups, resembling a universal prompting. In general, a threshold of 0.4 offers the highest adaptability across diverse semantic shifts. Further sensitivity analyses for other hyperparameters are presented in Appendix \ref{app_sec:add_param_sensitivity}.

\begin{figure}[t!]
\hspace*{-0.1cm}  
\includegraphics[width=1.00\columnwidth]{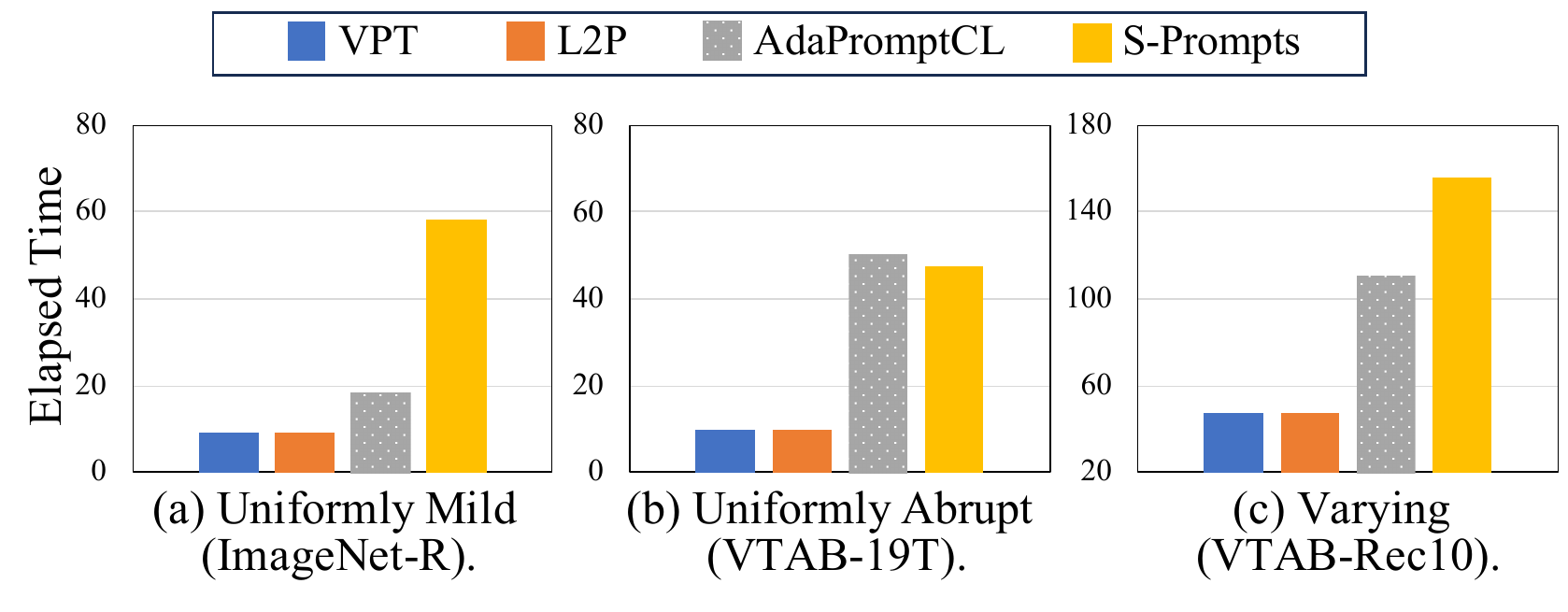}
\vspace*{-0.79cm}
\caption{Elapsed GPU runtime\,(mins) of prompt tuning-based methods across three shift scenarios.}
\vspace*{-0.1cm}
\label{fig:complexity}
\end{figure}

\subsection{Computational Complexity}

Figure \ref{fig:complexity} presents the elapsed GPU runtime for different prompting methods under three shift scenarios. The result indicates that the GPU runtime for \algname{} is contingent upon the severity of the semantic shifts. In scenarios with uniformly mild shifts, the runtime for \algname{} is comparable to that of universal prompting methods whereas in scenarios with abrupt semantic shifts, the runtime aligns with specific prompting methods. \algname{} incurs a slightly increased runtime in comparison to both universal and specific methods, which is attributed to the warm-up phase to extract semantic representations in Eq.~\eqref{eq:task_semantic_embedding}.

\subsection{Detailed Analyses}
\noindent\textbf{Generalization across Different Backbone Scales.}
Table \ref{tbl:gen_over_backbones} presents a comparison of \algname{} against baseline CL methods on varying semantic shift scenarios, using differently scaled pretrained backbones\,\cite{dosovitskiy2021an}, specifically ViT-S/16 and ViT-T/16. Overall, this result demonstrates the challenge of adjusting to semantic shifts across variously scaled pretrained backbone architectures, while highlighting the broad generalizability of AdaPromptCL regardless of the backbone architecture. Notably, \algname{} achieves an average performance improvement of 17.58\% and 16.06\% over the baselines on ViT-S/16 and ViT-T/16, respectively.

\def\arraystretch{1.15}
\begin{table}[t!]
\small
\centering
\caption{Performance\,(last accuracy) comparison of AdaPromptCL against CL baselines on varying shift scenarios with two new differently scaled pretrained backbones: ViT-S/16 and ViT-T/16. The best values are highlighted in bold.
}
\vspace*{0.3cm}
\resizebox{0.99\linewidth}{!}{%
\begin{tabular}[c]
{@{}cc|ccccc|cc@{}}
\toprule
\multicolumn{1}{c}{{ {} }} 
& \multicolumn{1}{c|}{{ {} }} 
& \multicolumn{5}{c|}{{ {Prompt Tuning CL Algorithms}}} 
& \multicolumn{1}{c}{{ { }}}\\
\multicolumn{1}{c}{{\hspace{-0.18cm}\textbf{Backbones}}\hspace{-0.58cm}}
& \multicolumn{1}{c|}{\hspace{-0.cm}{Datasets}\hspace{-0.1cm}}
& \multicolumn{1}{c}{{L2P}}
& \multicolumn{1}{c}{{VPT}}
& \multicolumn{1}{c}{{LAE}}
& \multicolumn{1}{c}{{DP}}
& \multicolumn{1}{c|}{{S-Prompts}}
& \multicolumn{1}{c}{\hspace{-0.1cm}{\textbf{\algname{}}}\hspace{-0.2cm}}
\\ \midrule

\multirow{6}{*}{\hspace{-0.0cm}{\parbox{1.5cm}{\centering\hspace{-0.58cm}\textbf{ViT-S/16}}}\hspace{-0.58cm}} 
& \multirow{2}{*}{\hspace{-0.1cm}\makecell[c]{VTAB-Rec2}\hspace{-0.1cm}}
& \begin{tabular}[c]{@{}c@{}}{ \hspace{-0.18cm}{50.30}\hspace{-0.18cm} }\end{tabular}
& \begin{tabular}[c]{@{}c@{}}{ \hspace{-0.18cm}{50.50}\hspace{-0.18cm} }\end{tabular}
& \begin{tabular}[c]{@{}c@{}}{ \hspace{-0.18cm}{46.58}\hspace{-0.18cm} }\end{tabular}
& \begin{tabular}[c]{@{}c@{}}{ \hspace{-0.18cm}{51.43}\hspace{-0.18cm} }\end{tabular}
& \begin{tabular}[c]{@{}c@{}}{ \hspace{-0.18cm}{48.65}\hspace{-0.08cm} }\end{tabular}
& \begin{tabular}[c]{@{}c@{}}{ \hspace{-0.18cm}{\textbf{54.10}}\hspace{-0.18cm} }\end{tabular}
\\
& & \begin{tabular}[c]{@{}c@{}}{ \hspace{-0.18cm}\,($\pm$0.46)\hspace{-0.15cm} }\end{tabular}
& \begin{tabular}[c]{@{}c@{}}{ \hspace{-0.18cm}\,($\pm$0.83)\hspace{-0.15cm} }\end{tabular}
& \begin{tabular}[c]{@{}c@{}}{ \hspace{-0.18cm}\,($\pm$0.22)\hspace{-0.15cm} }\end{tabular}
& \begin{tabular}[c]{@{}c@{}}{ \hspace{-0.18cm}\,($\pm$0.17)\hspace{-0.15cm} }\end{tabular}
& \begin{tabular}[c]{@{}c@{}}{ \hspace{-0.18cm}\,($\pm$0.31)\hspace{-0.15cm} }\end{tabular}
& \begin{tabular}[c]{@{}c@{}}{ \hspace{-0.18cm}\,({$\pm$0.42})\hspace{-0.15cm} }\end{tabular}
\\ \addlinespace[0.8ex]
& \multirow{2}{*}{\hspace{-0.1cm}\makecell[c]{VTAB-Rec5}\hspace{-0.1cm}}
& \begin{tabular}[c]{@{}c@{}}{ \hspace{-0.18cm}{49.90}\hspace{-0.18cm} }\end{tabular}
& \begin{tabular}[c]{@{}c@{}}{ \hspace{-0.18cm}{52.43}\hspace{-0.18cm} }\end{tabular}
& \begin{tabular}[c]{@{}c@{}}{ \hspace{-0.18cm}{43.57}\hspace{-0.18cm} }\end{tabular}
& \begin{tabular}[c]{@{}c@{}}{ \hspace{-0.18cm}{52.67}\hspace{-0.18cm} }\end{tabular}
& \begin{tabular}[c]{@{}c@{}}{ \hspace{-0.18cm}{48.63}\hspace{-0.08cm} }\end{tabular}
& \begin{tabular}[c]{@{}c@{}}{ \hspace{-0.18cm}{\textbf{58.30}}\hspace{-0.18cm} }\end{tabular}
\\ 
& & \begin{tabular}[c]{@{}c@{}}{ \hspace{-0.18cm}\,($\pm$0.25)\hspace{-0.15cm} }\end{tabular}
& \begin{tabular}[c]{@{}c@{}}{ \hspace{-0.18cm}\,($\pm$0.91)\hspace{-0.15cm} }\end{tabular}
& \begin{tabular}[c]{@{}c@{}}{ \hspace{-0.18cm}\,($\pm$0.72)\hspace{-0.15cm} }\end{tabular}
& \begin{tabular}[c]{@{}c@{}}{ \hspace{-0.18cm}\,($\pm$2.24)\hspace{-0.15cm} }\end{tabular}
& \begin{tabular}[c]{@{}c@{}}{ \hspace{-0.18cm}\,($\pm$0.59)\hspace{-0.15cm} }\end{tabular}
& \begin{tabular}[c]{@{}c@{}}{ \hspace{-0.18cm}\,({$\pm$3.03})\hspace{-0.15cm} }\end{tabular}
\\  \addlinespace[0.8ex]
& \multirow{2}{*}{\hspace{-0.1cm}\makecell[c]{VTAB-Rec10}\hspace{-0.1cm}}
& \begin{tabular}[c]{@{}c@{}}{ \hspace{-0.18cm}{41.65}\hspace{-0.18cm} }\end{tabular}
& \begin{tabular}[c]{@{}c@{}}{ \hspace{-0.18cm}{51.35}\hspace{-0.18cm} }\end{tabular}
& \begin{tabular}[c]{@{}c@{}}{ \hspace{-0.18cm}{35.35}\hspace{-0.18cm} }\end{tabular}
& \begin{tabular}[c]{@{}c@{}}{ \hspace{-0.18cm}{51.10}\hspace{-0.18cm} }\end{tabular}
& \begin{tabular}[c]{@{}c@{}}{ \hspace{-0.18cm}{47.60}\hspace{-0.08cm} }\end{tabular}
& \begin{tabular}[c]{@{}c@{}}{ \hspace{-0.18cm}{\textbf{57.00}}\hspace{-0.18cm} }\end{tabular}
\\
& & \begin{tabular}[c]{@{}c@{}}{ \hspace{-0.18cm}\,($\pm$0.67)\hspace{-0.15cm} }\end{tabular}
& \begin{tabular}[c]{@{}c@{}}{ \hspace{-0.18cm}\,($\pm$3.50)\hspace{-0.15cm} }\end{tabular}
& \begin{tabular}[c]{@{}c@{}}{ \hspace{-0.18cm}\,($\pm$1.24)\hspace{-0.15cm} }\end{tabular}
& \begin{tabular}[c]{@{}c@{}}{ \hspace{-0.18cm}\,($\pm$3.32)\hspace{-0.15cm} }\end{tabular}
& \begin{tabular}[c]{@{}c@{}}{ \hspace{-0.18cm}\,($\pm$0.78)\hspace{-0.15cm} }\end{tabular}
& \begin{tabular}[c]{@{}c@{}}{ \hspace{-0.18cm}\,({$\pm$3.53})\hspace{-0.15cm} }\end{tabular}
\\ \addlinespace[0.8ex]
\midrule
\multirow{6}{*}{\hspace{-0.0cm}{\parbox{1.5cm}{\centering\hspace{-0.58cm}\textbf{ViT-T/16}}}\hspace{-0.58cm}} 
& \multirow{2}{*}{\hspace{-0.1cm}\makecell[c]{VTAB-Rec2}\hspace{-0.1cm}}
& \begin{tabular}[c]{@{}c@{}}{ \hspace{-0.18cm}{36.90}\hspace{-0.18cm} }\end{tabular}
& \begin{tabular}[c]{@{}c@{}}{ \hspace{-0.18cm}{36.34}\hspace{-0.18cm} }\end{tabular}
& \begin{tabular}[c]{@{}c@{}}{ \hspace{-0.18cm}{33.60}\hspace{-0.18cm} }\end{tabular}
& \begin{tabular}[c]{@{}c@{}}{ \hspace{-0.18cm}{37.30}\hspace{-0.18cm} }\end{tabular}
& \begin{tabular}[c]{@{}c@{}}{ \hspace{-0.18cm}{37.68}\hspace{-0.08cm} }\end{tabular}
& \begin{tabular}[c]{@{}c@{}}{ \hspace{-0.18cm}{\textbf{39.62}}\hspace{-0.18cm} }\end{tabular}
\\ 
& & \begin{tabular}[c]{@{}c@{}}{ \hspace{-0.18cm}\,($\pm$0.22)\hspace{-0.15cm} }\end{tabular}
& \begin{tabular}[c]{@{}c@{}}{ \hspace{-0.18cm}\,($\pm$0.40)\hspace{-0.15cm} }\end{tabular}
& \begin{tabular}[c]{@{}c@{}}{ \hspace{-0.18cm}\,($\pm$0.60)\hspace{-0.15cm} }\end{tabular}
& \begin{tabular}[c]{@{}c@{}}{ \hspace{-0.18cm}\,($\pm$0.25)\hspace{-0.15cm} }\end{tabular}
& \begin{tabular}[c]{@{}c@{}}{ \hspace{-0.18cm}\,($\pm$0.78)\hspace{-0.15cm} }\end{tabular}
& \begin{tabular}[c]{@{}c@{}}{ \hspace{-0.18cm}\,({$\pm$0.41})\hspace{-0.15cm} }\end{tabular}
\\ \addlinespace[0.8ex]
& \multirow{2}{*}{\hspace{-0.1cm}\makecell[c]{VTAB-Rec5}\hspace{-0.1cm}}
& \begin{tabular}[c]{@{}c@{}}{ \hspace{-0.18cm}{37.03}\hspace{-0.18cm} }\end{tabular}
& \begin{tabular}[c]{@{}c@{}}{ \hspace{-0.18cm}{34.77}\hspace{-0.18cm} }\end{tabular}
& \begin{tabular}[c]{@{}c@{}}{ \hspace{-0.18cm}{34.10}\hspace{-0.18cm} }\end{tabular}
& \begin{tabular}[c]{@{}c@{}}{ \hspace{-0.18cm}{36.67}\hspace{-0.18cm} }\end{tabular}
& \begin{tabular}[c]{@{}c@{}}{ \hspace{-0.18cm}{35.90}\hspace{-0.08cm} }\end{tabular}
& \begin{tabular}[c]{@{}c@{}}{ \hspace{-0.18cm}{\textbf{40.13}}\hspace{-0.18cm} }\end{tabular}
\\ 
& & \begin{tabular}[c]{@{}c@{}}{ \hspace{-0.18cm}\,($\pm$0.66)\hspace{-0.15cm} }\end{tabular}
& \begin{tabular}[c]{@{}c@{}}{ \hspace{-0.18cm}\,($\pm$0.71)\hspace{-0.15cm} }\end{tabular}
& \begin{tabular}[c]{@{}c@{}}{ \hspace{-0.18cm}\,($\pm$0.15)\hspace{-0.15cm} }\end{tabular}
& \begin{tabular}[c]{@{}c@{}}{ \hspace{-0.18cm}\,($\pm$0.65)\hspace{-0.15cm} }\end{tabular}
& \begin{tabular}[c]{@{}c@{}}{ \hspace{-0.18cm}\,($\pm$0.22)\hspace{-0.15cm} }\end{tabular}
& \begin{tabular}[c]{@{}c@{}}{ \hspace{-0.18cm}\,({$\pm$0.52})\hspace{-0.15cm} }\end{tabular}
\\  \addlinespace[0.8ex]
& \multirow{2}{*}{\hspace{-0.1cm}\makecell[c]{VTAB-Rec10}\hspace{-0.1cm}}
& \begin{tabular}[c]{@{}c@{}}{ \hspace{-0.18cm}{33.10}\hspace{-0.18cm} }\end{tabular}
& \begin{tabular}[c]{@{}c@{}}{ \hspace{-0.18cm}{31.37}\hspace{-0.18cm} }\end{tabular}
& \begin{tabular}[c]{@{}c@{}}{ \hspace{-0.18cm}{29.27}\hspace{-0.18cm} }\end{tabular}
& \begin{tabular}[c]{@{}c@{}}{ \hspace{-0.18cm}{32.30}\hspace{-0.18cm} }\end{tabular}
& \begin{tabular}[c]{@{}c@{}}{ \hspace{-0.18cm}{33.87}\hspace{-0.08cm} }\end{tabular}
& \begin{tabular}[c]{@{}c@{}}{ \hspace{-0.18cm}{\textbf{40.80}}\hspace{-0.18cm} }\end{tabular}
\\ 
& & \begin{tabular}[c]{@{}c@{}}{ \hspace{-0.18cm}\,($\pm$0.87)\hspace{-0.15cm} }\end{tabular}
& \begin{tabular}[c]{@{}c@{}}{ \hspace{-0.18cm}\,($\pm$0.77)\hspace{-0.15cm} }\end{tabular}
& \begin{tabular}[c]{@{}c@{}}{ \hspace{-0.18cm}\,($\pm$0.59)\hspace{-0.15cm} }\end{tabular}
& \begin{tabular}[c]{@{}c@{}}{ \hspace{-0.18cm}\,($\pm$0.74)\hspace{-0.15cm} }\end{tabular}
& \begin{tabular}[c]{@{}c@{}}{ \hspace{-0.18cm}\,($\pm$0.43)\hspace{-0.15cm} }\end{tabular}
& \begin{tabular}[c]{@{}c@{}}{ \hspace{-0.18cm}\,({$\pm$0.43})\hspace{-0.15cm} }\end{tabular}
\\
\bottomrule
\end{tabular}
}
\label{tbl:gen_over_backbones}
\end{table}

\begin{figure}[t!]
\hspace*{0.0cm} 
\includegraphics[width=1.00\columnwidth]{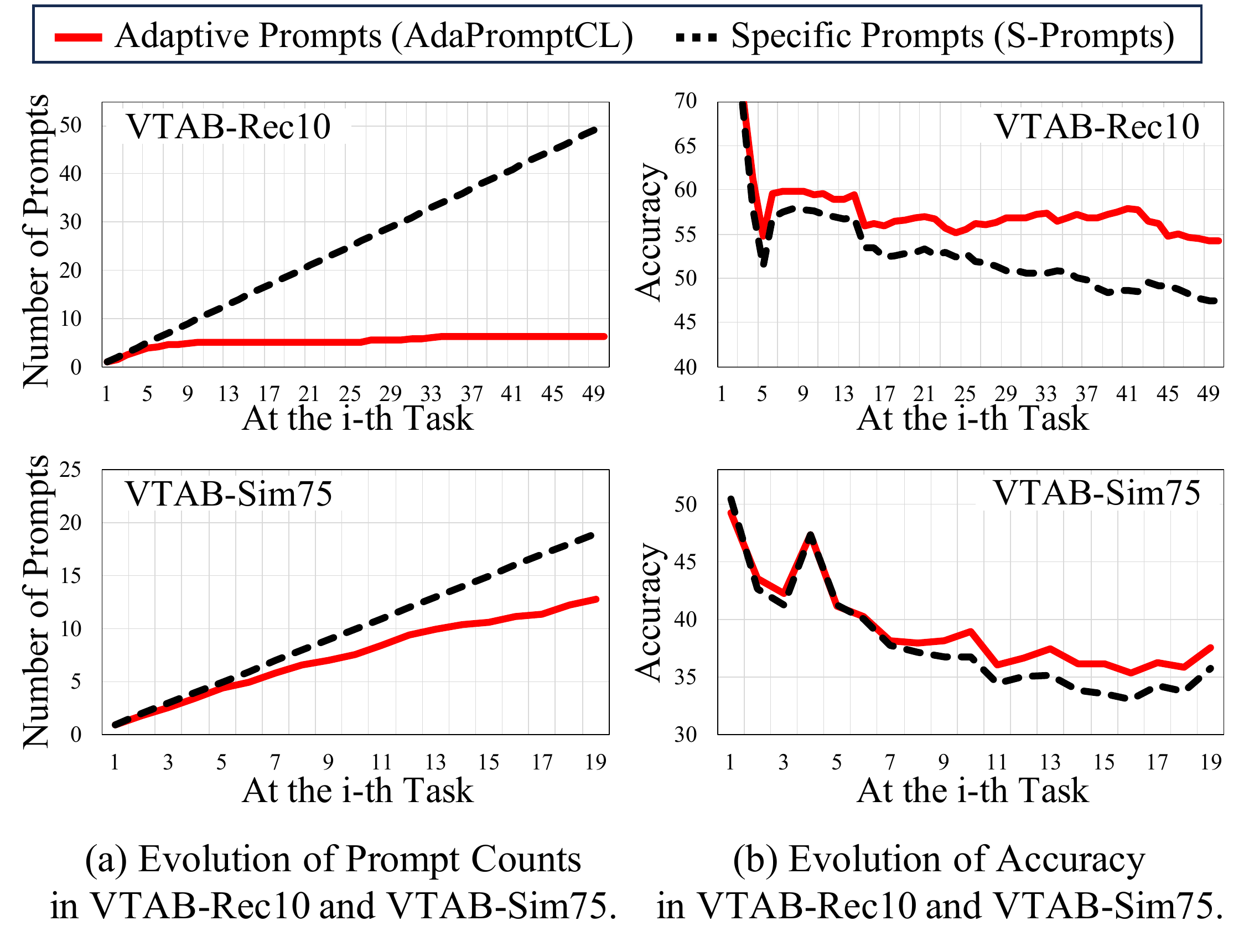}
\vspace*{-0.7cm}
\caption{A comparative analysis of the evolution of prompt counts and accuracy between specific\,(S-Prompts) and adaptive\,(\algname{}) prompting methods over CL streams based on VTAB-Rec10 and VTAB-Sim75.
}
\vspace*{-0.4cm}
\label{fig:trajectory_adaptive_prompting}
\end{figure}

\smallskip
\noindent\textbf{Dynamics of Prompting Approaches over CL Streams.} 
Figure \ref{fig:trajectory_adaptive_prompting} further presents a comparative analysis of the evolution in prompt counts and accuracy over task streams, contrasting adaptive prompting by \algname{} and specific prompting such as S-Prompts. In Figure \ref{fig:trajectory_adaptive_prompting}(a), the result shows AdaPromptCL's capability to dynamically adjust prompts according to task semantics, unlike specific prompting which consistently adds new prompts for each task. Additionally, Figure \ref{fig:trajectory_adaptive_prompting}(b) highlights a widening performance gap between AdaPromptCL and S-Prompts as CL streams advance. This advantage is attributed to the positive transfer among the tasks with high semantic similarities in \algname{}. Consequently, on the VTAB-Rec10, \algname{} realizes a 13.62\% performance improvement over S-Prompts while employing merely 12.40\% of the number of prompts that S-Prompts employs.

\section{Conclusion}
\label{sec:conclusion}

We propose \algname{} which exploits \emph{adaptive prompting} to address the diverse and unpredictable semantic shifts. The \emph{assign-and-refine semantic grouping} ensures that prompts are not only minimal but also precisely tailored to the evolving semantics of sequential tasks. 
Our results demonstrate the superior adaptability of \algname{} in managing various semantic shifts. Overall, we believe that our work sheds light on the importance of versatile and adaptable CL models for diverse real-world CL scenarios.

\section*{Impact Statement}
The deployment of \algname{}, an adaptive prompting approach, is of substantial relevance to continual learning\,(CL) scenarios in real-world settings, where data and conditions are subject to unpredictable changes, such as those in autonomous systems and dynamic financial markets. \algname{}'s effective handling of varying semantic shifts augments the robustness and utility of CL systems, which could lead to AI deployments that are both more reliable and efficient.

From an ethical standpoint, the use of semantic grouping within \algname{} necessitates caution as it could unintentionally reveal hidden biases among separated semantic groups. Consequently, it is important to diligently ensure the avoidance of unintentional prejudice or discrimination across different tasks or datasets, thus maintaining ethical integrity in the application of algorithmic systems.



\section*{Acknowledgements}
This work was supported by Institute of Information \& Communications Technology Planning \& Evaluation\,(IITP) grant funded by the Korea government\,(MSIT) (No.\ 2020-0-00862, DB4DL: High-Usability and Performance In-Memory Distributed DBMS for Deep Learning, 50\% and No.\ 2022-0-00157, Robust, Fair, Extensible Data-Centric Continual Learning, 50\%).

\bibliography{example_paper}
\bibliographystyle{icml2024}

\newpage
\appendix
\onecolumn
\section{Pseudocode of \algname{}}
\label{app_sec:overall_pseudocode}

The training procedure of \algname{} is described in Algorithm \ref{alg:overall}.
Our \textit{assign-and-refine} semantic grouping is executed through the semantic assignment\,(Lines 6--13) and the semantic refinement\,(Lines 14--19).
During the semantic assignment, \algname{} assigns the incoming task to an appropriate semantic group or initiates a new group\,(Lines 7--9). Concurrently, it updates the neighboring semantic groups and collects the prospective semantic groups using neighboring semantic groups\,(Lines 9--10) where the detailed procedure of collecting the prospective semantic groups is described in Algorithm \ref{alg:prep_groups}. In the semantic refinement, \algname{} finds the task order that yields the fewest semantic groups belonging to a neighboring semantic group updated by an incoming task\,(Line 15). If the semantic groups can be reduced, \algname{} refines the current semantic groups into the reduced groups and retrieves corresponding prompts from prospective semantic groups for these refined groups\,(Lines 16--18). Lastly, prompts and keys are initialized for each prospective semantic group, optimized with the classifier, and stored for future use\,(Lines 21--23). The process iterates for each task, progressively refining the semantic groups and tuning prompts.

\newlength{\oldtextfloatsep}
\setlength{\oldtextfloatsep}{\textfloatsep}
\setlength{\textfloatsep}{10pt} 
\newcommand{\INLINECOM}[1]{\STATE \textcolor{blue}{\textit{// #1 }}}

\begin{algorithm}[h!]
\footnotesize
\caption{\algname{}: \textit{Assign-and-refine} CL Framework}
\label{alg:overall}
\begin{algorithmic}[1]
\INPUT a sequence of tasks $\mathcal{T}$, data stream $\{D^1, \dots, D^\mathcal{T}\}$, threshold $R$, scaling factor $\gamma$, classifier $\phi$
\OUTPUT semantic groups $\mathcal{G}^{|\mathcal{T}|}$, prompts $\mathcal{P}$, keys $\mathcal{K}$, classifier $\phi$

\STATE {$\mathcal{G}^0 \leftarrow {\rm InitializeSemanticGroups}()$}
\STATE {$\hat{\mathcal{G}} \leftarrow {\rm InitializeProspectiveSemanticGroups}()$}
\STATE {$\mathcal{N}^0 \leftarrow {\rm InitializeNeighboringSemanticGroups}()$}
\STATE {$\mathcal{P}, \mathcal{K} \leftarrow {\rm PromptsRepository}(), {\rm KeysRepository}()$}

\FOR{each $\tau^t$ \textbf{in} $T$}
\STATE \textcolor{blue}{\textit{/* \textsc{Semantic Assignment: Section \ref{subsec:mac}} */}}
\label{lst:line:mac_s}
\STATE {$\hat{\rm P}^{\tau^t} \leftarrow {\rm WarmupPrompt}(D^t)$}
\label{lst:line:assign_s}
\STATE {$s(\tau^t) \leftarrow {\rm ExtractSemantic}(\hat{\rm P}^{\tau^t})$}
\STATE {${\mathcal{G}^t} \leftarrow {\rm AssignSemanticGroups}(s(\tau^t),{\mathcal{G}^{t-1}},R)$} 
\STATE {${\mathcal{N}^t} \leftarrow {\rm AssignNeighboringSemanticGroups}(s(\tau^t),{\mathcal{N}^{t-1}},\gamma R)$} 
\label{lst:line:assign_e}

\FOR{each ${{N}_i^t}$ \textbf{in} ${\mathcal{N}^t}$}
\IF{${\rm IsUpdated}({N}_i^t)$} 
\STATE {$\hat{\mathcal{G}} \leftarrow {\rm CollectProspectiveSemanticGroups}({N}_i^t, \hat{\mathcal{G}}, R)$}
\label{lst:line:prep}
\label{lst:line:mac_e}
\STATE \textcolor{blue}{\textit{/* \textsc{Semantic Refinement: Section \ref{subsec:mic}} */}}
\label{lst:line:mic_s}
\STATE {$\Bar{N}^t \leftarrow {\rm FindMininumGroupOrder}({N}_i^t,R)$}
\label{lst:line:ref_s}
\IF{${\rm IsReduced}(\Bar{N}_i^t, N_i^t)$} 
\STATE {${\rm RefineGroups}(\Bar{N}_i^t, N_i^t)$}
\STATE {${\rm RetrievePrompts}(\Bar{N}_i^t, \mathcal{P}, \mathcal{K})$}
\label{lst:line:ref_e}
\ENDIF
\label{lst:line:mic_e}
\STATE \textcolor{blue}{\textit{/* \textsc{Semantic Group Based Prompt Tuning: Section \ref{subsec:tr_inf}} */}}
\label{lst:line:train_s}
\STATE {$\theta_t \leftarrow {\rm InitializePromptsKeys}(\tau^t,\hat{\mathcal{G}})$}
\STATE {$\theta_t,\phi \leftarrow {\rm Train}(\theta_t,\phi,D^t)$}
\STATE {$\mathcal{P}, \mathcal{K} \leftarrow {\rm UpdatePromptsKeysRepositories}(\theta_t)$}
\label{lst:line:train_e}

\ENDIF
\ENDFOR
\ENDFOR
\STATE \textbf{return} $\mathcal{G}^{|\mathcal{T}|}$, $\mathcal{P}$, $\mathcal{K}$, $\phi$;
\end{algorithmic}
\end{algorithm}

\section{Prospective Prompts Collection}
\label{app_sec:prop_pseudocode}
{\noindent\textbf{Implementation Details.}}
This section provides an in-depth explanation of the \emph{prospective prompts} collection process, which was previously omitted due to space constraints. Prospective prompts are instrumental in ensuring that semantic groups are updated with prompts that are tailored to all the tasks they encompass. Thus, finding prospective prompts requires finding semantic groups that likely appear in the future by semantic refinement. Algorithm \ref{alg:prep_groups} outlines the process for collecting prospective semantic groups, given a neighboring semantic group and the preceding prospective semantic groups as indicated in Line 13 of Algorithm \ref{alg:overall}.

Overall, as suggested in \cite{sil_clusters}, the first step is to ascertain the ideal number of clusters based on task semantics, using the silhouette score\,\cite{rousseeuw1987silhouettes}. This is followed by the collection of prospective semantic groups with the \textit{k}-means clustering method.
The algorithm commences by initializing a storage for silhouette scores\,(Line 1).  It then iteratively computes silhouette scores for different cluster configurations and records them\,(Lines 3--8). Subsequently, it determines the top-$\eta$ optimal cluster counts that maximize the silhouette score\,(Line 9). With these optimal counts, the \textit{k}-means algorithm is applied (Lines 11--15), and the newly identified candidate groups are amalgamated with existing prospective semantic groups\,(Line 13). For our experiments, we set the clustering iteration count $r$ to 100 and the number of representative clusters counts $\eta$ to 2. Please see Appendix \ref{app_sec:add_param_sensitivity} for sensitivity analyses regarding these hyperparameters.
 
\setlength{\oldtextfloatsep}{\textfloatsep}
\setlength{\textfloatsep}{10pt} 

\begin{algorithm}[t!]
\footnotesize
\caption{Prospective Semantic Groups Collection}
\label{alg:prep_groups}
\begin{algorithmic}[1]
\INPUT neighboring semantic group ${N}_i^t$, preceding prospective semantic groups $\hat{\mathcal{G}}$, clustering threshold $R$, clustering iteration count $r$, number of representative clusters counts $\eta$

\OUTPUT prospective semantic groups $\hat{\mathcal{G}}$
\STATE {$\mathcal{S} \leftarrow {\rm InitializeSilhouetteMemory}()$}
\label{lst:line:init_s}
\label{lst:line:init_e}
\STATE \textcolor{blue}{\textit{/* \textsc{Identification of Representative Cluster Counts via Silhouette Score} */}}
\FOR{each ${j}$ {\bf from} 1 {\bf to} $r$}
\label{lst:line:sil_s}
\STATE {${L} \leftarrow {{\rm LabelClusters}({\rm Cluster}(N_i^t, R))}$}
\STATE {${k} \leftarrow {{\rm CountClusters}(L)}$}
\STATE {${s} \leftarrow {{\rm MeasureSilhouetteScore}(N_i^t, L)}$}
\STATE {$\mathcal{S} \leftarrow {{\rm UpdateSilhouetteMemory}(\mathcal{S},\{(k, s)\}) } $}
\label{lst:line:sil_e}
\ENDFOR
\STATE {${K} \leftarrow \arg\max_{K:\ |K|\leq \eta} \sum_{k \in K}{\rm AverageSilhouette}(\mathcal{S},k)$}
\label{lst:line:cnt}
\STATE \textcolor{blue}{\textit{/* \textsc{Collection of Prospective Semantic Groups via Representative Cluster Counts} */}}
\label{lst:line:add_s}
\FOR{each ${j}$ {\bf from} 1 {\bf to} $r$}
\FOR{each $k$ \textbf{in} $K$}
\STATE {${\hat{\mathcal{G}}} \leftarrow \hat{\mathcal{G}} \cup {\text{\emph{k}-means}(N_i^t, k)}$}
\label{lst:line:add_e}
\ENDFOR
\ENDFOR
\STATE \textbf{return} $\hat{\mathcal{G}}^t$;

\end{algorithmic}
\end{algorithm}

\smallskip
{\noindent\textbf{Computational Complexity.}}
Table \ref{tbl:prep_complexity} compares GPU memory usage and runtime between \algname{} configurations with and without the incorporation of prospective semantic groups. The `No Refine' variant, described in Section 5.3, denotes the configuration that does not include prospective semantic groups, thus lacking the capability for semantic group refinement. In general, computational demand is generally influenced by the extent of semantic shift within the tasks. uniformly mild and abrupt scenarios do not typically benefit from prospective semantic groups due to the unambiguous nature of task relationships. In the varying shifting scenario, however, the varying degrees of semantic shifts between tasks increase the demand for prospective semantic groups. Specifically, an average of 95.2 prospective semantic groups are utilized. Despite this, the increase in total GPU memory usage is marginal, at 149.6MB\,(1.6\%), attributable to the small memory footprint of the prompts. Moreover, this leads to a mere 14 minutes\,(12.5\%) increase in GPU running time, as prospective semantic groups necessitate a short period of fine-tuning for alignment with the current task.
\newcolumntype{C}[1]{>{\centering\arraybackslash}p{#1}}
\newcolumntype{L}[1]{>{\raggedright\arraybackslash}p{#1}}
\def\arraystretch{1.1}
\begin{table}[h]
\centering
\caption{Computational overhead incurred by prospective semantic groups of \algname{} across diverse semantic shift scenarios, presenting GPU memory\,(in MB) and running time\,(in minutes). All results are obtained utilizing automatic mixed precision to optimize runtime and memory consumption.}
\vspace*{0.4cm}
\resizebox{0.75\columnwidth}{!}{%
\begin{tabular}[c]
{@{}c|ccc|ccc|ccc@{}}
\toprule
\multicolumn{1}{c|}{{Scenarios}\hspace{-0.4cm}} & \multicolumn{3}{c}{\textbf{Uniformly Mild}\,(ImageNet-R)} & \multicolumn{3}{c}{\textbf{Uniformly Abrupt}\,(VTAB-19T)} & \multicolumn{3}{c}{\textbf{Varying}\,(VTAB-Rec10)}  \\
\addlinespace[1.50ex]
\addlinespace[-0.70ex]\multirow{2}{*}{\hspace{-0.0cm}{{Metrics}}\hspace{-0.01cm}} 
& \begin{tabular}[c]{@{}c@{}}{ \scalebox{0.80}{\textbf{\#Prospective}} }\end{tabular} 
& \begin{tabular}[c]{@{}c@{}}{ \scalebox{0.80}{\textbf{GPU}} }\end{tabular} 
& \begin{tabular}[c]{@{}c@{}}{ \scalebox{0.80}{\textbf{Running}} }\end{tabular}
& \begin{tabular}[c]{@{}c@{}}{ \scalebox{0.80}{\textbf{\#Prospective}} }\end{tabular} 
& \begin{tabular}[c]{@{}c@{}}{ \scalebox{0.80}{\textbf{GPU}} }\end{tabular} 
& \begin{tabular}[c]{@{}c@{}}{ \scalebox{0.80}{\textbf{Running}} }\end{tabular}
& \begin{tabular}[c]{@{}c@{}}{ \scalebox{0.80}{\textbf{\#Prospective}} }\end{tabular} 
& \begin{tabular}[c]{@{}c@{}}{ \scalebox{0.80}{\textbf{GPU}} }\end{tabular} 
& \begin{tabular}[c]{@{}c@{}}{ \scalebox{0.80}{\textbf{Running}} }\end{tabular}
\\ \addlinespace[-0.70ex]
& \begin{tabular}[c]{@{}c@{}}{ \scalebox{0.80}{\textbf{Prompts}} }\end{tabular} 
& \begin{tabular}[c]{@{}c@{}}{ \scalebox{0.80}{\textbf{Memory}} }\end{tabular} 
& \begin{tabular}[c]{@{}c@{}}{ \scalebox{0.80}{\textbf{Time}} }\end{tabular}
& \begin{tabular}[c]{@{}c@{}}{ \scalebox{0.80}{\textbf{Prompts}} }\end{tabular} 
& \begin{tabular}[c]{@{}c@{}}{ \scalebox{0.80}{\textbf{Memory}} }\end{tabular} 
& \begin{tabular}[c]{@{}c@{}}{ \scalebox{0.80}{\textbf{Time}} }\end{tabular}
& \begin{tabular}[c]{@{}c@{}}{ \scalebox{0.80}{\textbf{Prompts}} }\end{tabular} 
& \begin{tabular}[c]{@{}c@{}}{ \scalebox{0.80}{\textbf{Memory}} }\end{tabular} 
& \begin{tabular}[c]{@{}c@{}}{ \scalebox{0.80}{\textbf{Time}} }\end{tabular}
\\
\addlinespace[0.70ex]\toprule

\multirow{2}{*}{\makecell[c]{\textbf{No Refine}}}
& \begin{tabular}[c]{@{}c@{}}{ \hspace{-0.01cm}{-}\hspace{-0.01cm} }\end{tabular}
& \begin{tabular}[c]{@{}c@{}}{ \hspace{-0.01cm}{8859.0}\hspace{-0.01cm} }\end{tabular}
& \begin{tabular}[c]{@{}c@{}}{ \hspace{-0.01cm}{{18.2}}\hspace{-0.01cm} }\end{tabular}
& \begin{tabular}[c]{@{}c@{}}{ \hspace{-0.01cm}{-}\hspace{-0.01cm} }\end{tabular}
& \begin{tabular}[c]{@{}c@{}}{ \hspace{-0.01cm}{9093.0}\hspace{-0.01cm} }\end{tabular}
& \begin{tabular}[c]{@{}c@{}}{ \hspace{-0.01cm}{{62.4}}\hspace{-0.01cm} }\end{tabular}
& \begin{tabular}[c]{@{}c@{}}{ \hspace{-0.01cm}{-}\hspace{-0.01cm} }\end{tabular}
& \begin{tabular}[c]{@{}c@{}}{ \hspace{-0.01cm}{9107.0}\hspace{-0.01cm} }\end{tabular}
& \begin{tabular}[c]{@{}c@{}}{ \hspace{-0.01cm}{112.0}\hspace{-0.01cm} }\end{tabular}
\\
& \begin{tabular}[c]{@{}c@{}}{ \hspace{-0.01cm}\scalebox{0.99}(-)\hspace{-0.01cm} }\end{tabular}
& \begin{tabular}[c]{@{}c@{}}{ \hspace{-0.01cm}\scalebox{0.99}($\pm$0.0)\hspace{-0.01cm} }\end{tabular}
& \begin{tabular}[c]{@{}c@{}}{ \hspace{-0.01cm}\scalebox{0.99}($\pm$0.2)\hspace{-0.01cm} }\end{tabular}
& \begin{tabular}[c]{@{}c@{}}{ \hspace{-0.01cm}\scalebox{0.99}(-)\hspace{-0.01cm} }\end{tabular}
& \begin{tabular}[c]{@{}c@{}}{ \hspace{-0.01cm}\scalebox{0.99}($\pm$0.0)\hspace{-0.01cm} }\end{tabular}
& \begin{tabular}[c]{@{}c@{}}{ \hspace{-0.01cm}\scalebox{0.99}({$\pm$0.2})\hspace{-0.01cm} }\end{tabular}
& \begin{tabular}[c]{@{}c@{}}{ \hspace{-0.01cm}\scalebox{0.99}(-)\hspace{-0.01cm} }\end{tabular}
& \begin{tabular}[c]{@{}c@{}}{ \hspace{-0.01cm}\scalebox{0.99}($\pm$0.0)\hspace{-0.01cm} }\end{tabular}
& \begin{tabular}[c]{@{}c@{}}{ \hspace{-0.01cm}\scalebox{0.99}({$\pm$2.3})\hspace{-0.01cm} }\end{tabular}
\\
\addlinespace[1.10ex]
\multirow{2}{*}{\makecell[c]{\textbf{\algname{}}}}
& \begin{tabular}[c]{@{}c@{}}{ \hspace{-0.01cm}{0.0}\hspace{-0.01cm} }\end{tabular}
& \begin{tabular}[c]{@{}c@{}}{ \hspace{-0.01cm}{8859.0}\hspace{-0.01cm} }\end{tabular}
& \begin{tabular}[c]{@{}c@{}}{ \hspace{-0.01cm}{{18.4}}\hspace{-0.01cm} }\end{tabular}
& \begin{tabular}[c]{@{}c@{}}{ \hspace{-0.01cm}{95.2}\hspace{-0.01cm} }\end{tabular}
& \begin{tabular}[c]{@{}c@{}}{ \hspace{-0.01cm}{9256.6}\hspace{-0.01cm} }\end{tabular}
& \begin{tabular}[c]{@{}c@{}}{ \hspace{-0.01cm}{{126.0}}\hspace{-0.01cm} }\end{tabular}
& \begin{tabular}[c]{@{}c@{}}{ \hspace{-0.01cm}{0.6}\hspace{-0.01cm} }\end{tabular}
& \begin{tabular}[c]{@{}c@{}}{ \hspace{-0.01cm}{9103.8}\hspace{-0.01cm} }\end{tabular}
& \begin{tabular}[c]{@{}c@{}}{ \hspace{-0.01cm}{{63.0}}\hspace{-0.01cm} }\end{tabular}
\\
& \begin{tabular}[c]{@{}c@{}}{ \hspace{-0.01cm}\scalebox{0.99}($\pm$0.0)\hspace{-0.01cm} }\end{tabular}
& \begin{tabular}[c]{@{}c@{}}{ \hspace{-0.01cm}\scalebox{0.99}($\pm$0.0)\hspace{-0.01cm} }\end{tabular}
& \begin{tabular}[c]{@{}c@{}}{ \hspace{-0.01cm}\scalebox{0.99}($\pm$0.2)\hspace{-0.01cm} }\end{tabular}
& \begin{tabular}[c]{@{}c@{}}{ \hspace{-0.01cm}\scalebox{0.99}($\pm$10.2)\hspace{-0.01cm} }\end{tabular}
& \begin{tabular}[c]{@{}c@{}}{ \hspace{-0.01cm}\scalebox{0.99}($\pm$8.2)\hspace{-0.01cm} }\end{tabular}
& \begin{tabular}[c]{@{}c@{}}{ \hspace{-0.01cm}\scalebox{0.99}({$\pm$4.1})\hspace{-0.01cm} }\end{tabular}
& \begin{tabular}[c]{@{}c@{}}{ \hspace{-0.01cm}\scalebox{0.99}($\pm$0.2)\hspace{-0.01cm} }\end{tabular}
& \begin{tabular}[c]{@{}c@{}}{ \hspace{-0.01cm}\scalebox{0.99}($\pm$4.4)\hspace{-0.01cm} }\end{tabular}
& \begin{tabular}[c]{@{}c@{}}{ \hspace{-0.01cm}\scalebox{0.99}({$\pm$0.3})\hspace{-0.01cm} }\end{tabular}
\\
\bottomrule
\end{tabular}
}
\label{tbl:prep_complexity}
\end{table}

\section{VTAB Datasets Preparation}
\label{app_sec:vtab_dataset}
Visual Task Adaptation Benchmark\,(VTAB)\,\cite{zhai2019large} is a dataset in the field of machine learning, specifically designed for evaluating the adaptability and generalization capabilities of visual representation learning methods. VTAB consists of 19 tasks from diverse domains such as natural images, specialized tasks (like medical imaging or satellite imagery), and structured tasks (such as object counting or predicting depth from images). This diversity ensures that models are tested on a wide range of visual understanding capabilities.

\begin{figure}[t!]
\begin{center}
\includegraphics[width=0.75\textwidth,]{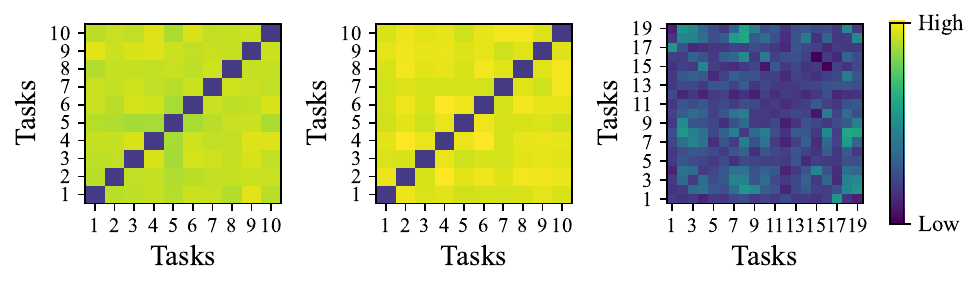}    
\end{center}

\hspace*{3.6cm} {\small (a) CIFAR100.} \hspace*{1.6cm} {\small (b) ImageNet-R.} \hspace*{1.9cm} {\small (c) VTAB.}
\caption{Task-to-task semantic similarity matrices for CL datasets: (a) CIFAR100, (b) ImageNet-R, and (c) VTAB, with similarity measured by cosine similarity. Lighter colors indicate higher similarity\,(`High'), while darker ones correspond to lower similarity\,(`Low').  }
\label{fig:vis_simil}
\end{figure}

\smallskip
{\noindent\textbf{Task Semantic Similarity Comparison.}}
Figure \ref{fig:vis_simil} shows that the degree of semantic similarity among tasks in VTAB is significantly low when compared to CIFAR100 and ImageNet-R. Semantic similarity is assessed by training task-specific prompts and extracting semantic representations following Eq.~\eqref{eq:task_semantic_embedding}. The matrices reveal a notable contrast in semantic similarities: CIFAR100 and ImageNet-R show significantly high similarities, reflecting mild semantic shifts, whereas the VTAB shows low similarities, indicating more abrupt semantic shifts between tasks.

\smallskip
{\noindent\textbf{Uniformly Abrupt Scenarios with VTAB.}}
By virtue of its wide-ranging domains, VTAB is employed to symbolize abrupt semantic shifts where every dataset is regarded as an individual task, referred to as VTAB-19T. Furthermore, we construct VTAB-5T, a collection of the five datasets that are most distinct in terms of semantics: \textit{(clevr-count}, \textit{resisc45}, \textit{diabetic-retinopathy}, \textit{oxford-flowers102}, \textit{dsprites-loc)}. 
In order to obtain these distinct datasets, task semantic representations are extracted for each task as in Eq.~\eqref{eq:task_semantic_embedding}, following the training of prompt for each individual task. Then, using spectral clustering\,\cite{ng2001spectral}, we cluster the semantic representations of 19 tasks into five distinct clusters and select a single task from each cluster, for a total of five datasets. 



\smallskip
{\noindent\textbf{Varying Scenarios with VTAB.}}
For varying semantic shifting scenarios, VTAB-SimS is designed to mimic the overlapping of tasks in VTAB, following the realistic and practical CL setup\,\cite{koh2022online}. This setup assumes that the model regularly encounters new classes while previously observed classes may reappear in subsequent tasks. Accordingly, VTAB datasets are divided into two categories: dissimilar datasets, introducing entirely new classes to the model, and \textit{similar} datasets, where tasks are comprised of a mix of classes from different datasets. Specifically, each overlapping task in VTAB-SimS includes S\% from other datasets and (100-S)\% of its dataset. For our experiments, we randomly designate 9 tasks as dissimilar and 10 as similar, experimenting with three different values for S: 25\%, 50\%, and 75\%. A higher S value renders tasks more semantically similar due to a greater proportion of shared classes. Meanwhile, VTAB-RecR provides an additional scenario of varying semantic shifting through the recurrence of tasks where a higher R increases similarities between tasks as semantically similar tasks recur more frequently. In VTAB-RecR, data instances for recurrent tasks are randomly drawn from their respective original tasks.


\section{Additional Experiment Results}
\label{app_sec:additional_exp}
\def\arraystretch{1.2}
\begin{table*}[t!]
\small
\centering
\caption{
Performance comparison of \algname{} against CL baselines using prompt tuning across uniformly and varying shift scenarios. This table shows both the forgetting for individual datasets and the average forgetting\,(presented in the final row), highlighting the relative performance degradation of each baseline in comparison to \algname{}.
}
\vspace*{0.2cm}
\resizebox{0.85\linewidth}{!}{%
\begin{tabular}[c]
{@{}cc|ccccc|cc@{}}
\toprule
\multirow{2}{*}{\hspace{-0.0cm}\makecell[c]{Shifting\\Scenarios}\hspace{-0.0cm}}&\multirow{2}{*}{\makecell[c]{CL\\Datasets}} 
& \multicolumn{5}{c|}{{ {Prompt Tuning CL Algorithms}}} 
&  \\ 
& & \multicolumn{1}{c}{\hspace{-0.15cm}{L2P}\hspace{-0.15cm}}
& \multicolumn{1}{c}{\hspace{-0.15cm}{VPT}\hspace{-0.15cm}}
& \multicolumn{1}{c}{\hspace{-0.15cm}{LAE}\hspace{-0.15cm}}
& \multicolumn{1}{c}{\hspace{-0.15cm}{DP}\hspace{-0.15cm}}
& \multicolumn{1}{c|}{\hspace{-0.15cm}{S-Prompts}\hspace{-0.15cm}}
& \multicolumn{1}{c}{\hspace{-0.05cm}{\textbf{\algname{}}}\hspace{-0.15cm}}
& 
\\ \midrule
\multirow{6}{*}{\hspace{-0.0cm}{\textbf{{Varying}}}\hspace{-0.15cm}} 
& \multirow{1}{*}{\makecell[c]{VTAB-Sim25}}
& \begin{tabular}[c]{@{}c@{}}{ \hspace{-0.15cm}{5.96}\scalebox{0.99}\,($\pm$0.48)\hspace{-0.15cm} }\end{tabular}
& \begin{tabular}[c]{@{}c@{}}{ \hspace{-0.15cm}{5.11}\scalebox{0.99}\,($\pm$0.38)\hspace{-0.15cm} }\end{tabular}
& \begin{tabular}[c]{@{}c@{}}{ \hspace{-0.15cm}{8.09}\scalebox{0.99}\,($\pm$0.54)\hspace{-0.15cm} }\end{tabular}
& \begin{tabular}[c]{@{}c@{}}{ \hspace{-0.15cm}{5.45}\scalebox{0.99}\,($\pm$0.41)\hspace{-0.15cm} }\end{tabular}
& \begin{tabular}[c]{@{}c@{}}{ \hspace{-0.15cm}{6.39}\scalebox{0.99}\,($\pm$0.49)\hspace{-0.15cm} }\end{tabular}
& \begin{tabular}[c]{@{}c@{}}{ \hspace{-0.15cm}{6.16}\scalebox{0.99}\,($\pm$0.55)\hspace{-0.15cm} }\end{tabular}
\\ 
& \multirow{1}{*}{\makecell[c]{VTAB-Sim50}}
& \begin{tabular}[c]{@{}c@{}}{ \hspace{-0.15cm}{4.94}\scalebox{0.99}\,($\pm$0.35)\hspace{-0.15cm} }\end{tabular}
& \begin{tabular}[c]{@{}c@{}}{ \hspace{-0.15cm}{5.04}\scalebox{0.99}\,($\pm$0.42)\hspace{-0.15cm} }\end{tabular}
& \begin{tabular}[c]{@{}c@{}}{ \hspace{-0.15cm}{6.60}\scalebox{0.99}\,($\pm$0.37)\hspace{-0.15cm} }\end{tabular}
& \begin{tabular}[c]{@{}c@{}}{ \hspace{-0.15cm}{5.01}\scalebox{0.99}\,($\pm$0.48)\hspace{-0.15cm} }\end{tabular}
& \begin{tabular}[c]{@{}c@{}}{ \hspace{-0.15cm}{5.43}\scalebox{0.99}\,($\pm$0.30)\hspace{-0.15cm} }\end{tabular}
& \begin{tabular}[c]{@{}c@{}}{ \hspace{-0.15cm}{4.34}\scalebox{0.99}\,($\pm$0.44)\hspace{-0.15cm} }\end{tabular}
\\ 
& \multirow{1}{*}{\makecell[c]{VTAB-Sim75}}
& \begin{tabular}[c]{@{}c@{}}{ \hspace{-0.15cm}{3.99}\scalebox{0.99}\,($\pm$0.26)\hspace{-0.15cm} }\end{tabular}
& \begin{tabular}[c]{@{}c@{}}{ \hspace{-0.15cm}{3.96}\scalebox{0.99}\,($\pm$0.27)\hspace{-0.15cm} }\end{tabular}
& \begin{tabular}[c]{@{}c@{}}{ \hspace{-0.15cm}{5.39}\scalebox{0.99}\,($\pm$0.30)\hspace{-0.15cm} }\end{tabular}
& \begin{tabular}[c]{@{}c@{}}{ \hspace{-0.15cm}{3.49}\scalebox{0.99}\,($\pm$0.26)\hspace{-0.15cm} }\end{tabular}
& \begin{tabular}[c]{@{}c@{}}{ \hspace{-0.15cm}{3.67}\scalebox{0.99}\,($\pm$0.20)\hspace{-0.15cm} }\end{tabular}
& \begin{tabular}[c]{@{}c@{}}{ \hspace{-0.15cm}{4.16}\scalebox{0.99}\,($\pm$0.41)\hspace{-0.15cm} }\end{tabular}
\\ \addlinespace[0.3ex]\cline{2-9}\addlinespace[0.5ex]
& \multirow{1}{*}{\makecell[c]{VTAB-Rec2}}
& \begin{tabular}[c]{@{}c@{}}{ \hspace{-0.15cm}{6.69}\scalebox{0.99}\,($\pm$0.67)\hspace{-0.15cm} }\end{tabular}
& \begin{tabular}[c]{@{}c@{}}{ \hspace{-0.15cm}{6.71}\scalebox{0.99}\,($\pm$0.65)\hspace{-0.15cm} }\end{tabular}
& \begin{tabular}[c]{@{}c@{}}{ \hspace{-0.15cm}{10.59}\scalebox{0.99}\,($\pm$0.56)\hspace{-0.15cm} }\end{tabular}
& \begin{tabular}[c]{@{}c@{}}{ \hspace{-0.15cm}{6.00}\scalebox{0.99}\,($\pm$0.69)\hspace{-0.15cm} }\end{tabular}
& \begin{tabular}[c]{@{}c@{}}{ \hspace{-0.15cm}{6.18}\scalebox{0.99}\,($\pm$0.60)\hspace{-0.15cm} }\end{tabular}
& \begin{tabular}[c]{@{}c@{}}{ \hspace{-0.15cm}{{5.63}}\scalebox{0.99}\,($\pm$0.57)\hspace{-0.15cm} }\end{tabular}
\\ 
& \multirow{1}{*}{\makecell[c]{VTAB-Rec5}}
& \begin{tabular}[c]{@{}c@{}}{ \hspace{-0.15cm}{2.82}\scalebox{0.99}\,($\pm$0.19)\hspace{-0.15cm} }\end{tabular}
& \begin{tabular}[c]{@{}c@{}}{ \hspace{-0.15cm}{3.63}\scalebox{0.99}\,($\pm$0.06)\hspace{-0.15cm} }\end{tabular}
& \begin{tabular}[c]{@{}c@{}}{ \hspace{-0.15cm}{4.97}\scalebox{0.99}\,($\pm$0.33)\hspace{-0.15cm} }\end{tabular}
& \begin{tabular}[c]{@{}c@{}}{ \hspace{-0.15cm}{3.53}\scalebox{0.99}\,($\pm$0.27)\hspace{-0.15cm} }\end{tabular}
& \begin{tabular}[c]{@{}c@{}}{ \hspace{-0.15cm}{4.27}\scalebox{0.99}\,($\pm$0.16)\hspace{-0.15cm} }\end{tabular}
& \begin{tabular}[c]{@{}c@{}}{ \hspace{-0.15cm}{2.71}\scalebox{0.99}\,($\pm$0.17)\hspace{-0.15cm} }\end{tabular}
\\ 
& \multirow{1}{*}{\makecell[c]{VTAB-Rec10}}
& \begin{tabular}[c]{@{}c@{}}{ \hspace{-0.15cm}{6.33}\scalebox{0.99}\,($\pm$0.37)\hspace{-0.15cm} }\end{tabular}
& \begin{tabular}[c]{@{}c@{}}{ \hspace{-0.15cm}{8.25}\scalebox{0.99}\,($\pm$0.53)\hspace{-0.15cm} }\end{tabular}
& \begin{tabular}[c]{@{}c@{}}{ \hspace{-0.15cm}{7.87}\scalebox{0.99}\,($\pm$0.43)\hspace{-0.15cm} }\end{tabular}
& \begin{tabular}[c]{@{}c@{}}{ \hspace{-0.15cm}{8.07}\scalebox{0.99}\,($\pm$0.46)\hspace{-0.15cm} }\end{tabular}
& \begin{tabular}[c]{@{}c@{}}{ \hspace{-0.15cm}{8.00}\scalebox{0.99}\,($\pm$0.52)\hspace{-0.15cm} }\end{tabular}
& \begin{tabular}[c]{@{}c@{}}{ \hspace{-0.15cm}{8.32}\scalebox{0.99}\,($\pm$0.62)\hspace{-0.15cm} }\end{tabular}
\\ \midrule
\multirow{2}{*}{\hspace{-0.0cm}{\parbox{1.5cm}{\centering{\textbf{Uniformly Mild}}}}\hspace{-0.15cm}} 
& \multirow{1}{*}{\makecell[c]{ImageNet-R}}
& \begin{tabular}[c]{@{}c@{}}{ \hspace{-0.15cm}{5.17}\scalebox{0.99}\,($\pm$0.10)\hspace{-0.15cm} }\end{tabular}
& \begin{tabular}[c]{@{}c@{}}{ \hspace{-0.15cm}{5.27}\scalebox{0.99}\,($\pm$0.08)\hspace{-0.15cm} }\end{tabular}
& \begin{tabular}[c]{@{}c@{}}{ \hspace{-0.15cm}{6.05}\scalebox{0.99}\,($\pm$0.21)\hspace{-0.15cm} }\end{tabular}
& \begin{tabular}[c]{@{}c@{}}{ \hspace{-0.15cm}{{4.99}}\scalebox{0.99}\,($\pm$0.18)\hspace{-0.15cm} }\end{tabular}
& \begin{tabular}[c]{@{}c@{}}{ \hspace{-0.15cm}{7.71}\scalebox{0.99}\,($\pm$0.22)\hspace{-0.15cm} }\end{tabular}
& \begin{tabular}[c]{@{}c@{}}{ \hspace{-0.15cm}{{4.94}}\scalebox{0.99}\,($\pm$0.08)\hspace{-0.15cm} }\end{tabular}
\\ 
& \multirow{1}{*}{\makecell[c]{CIFAR100}}
& \begin{tabular}[c]{@{}c@{}}{ \hspace{-0.15cm}{{5.59}}\scalebox{0.99}\,($\pm$0.08)\hspace{-0.15cm} }\end{tabular}
& \begin{tabular}[c]{@{}c@{}}{ \hspace{-0.15cm}{6.19}\scalebox{0.99}\,($\pm$0.07)\hspace{-0.15cm} }\end{tabular}
& \begin{tabular}[c]{@{}c@{}}{ \hspace{-0.15cm}{6.05}\scalebox{0.99}\,($\pm$0.21)\hspace{-0.15cm} }\end{tabular}
& \begin{tabular}[c]{@{}c@{}}{ \hspace{-0.15cm}{{5.04}}\scalebox{0.99}\,($\pm$0.15)\hspace{-0.15cm} }\end{tabular}
& \begin{tabular}[c]{@{}c@{}}{ \hspace{-0.15cm}{6.41}\scalebox{0.99}\,($\pm$0.13)\hspace{-0.15cm} }\end{tabular}
& \begin{tabular}[c]{@{}c@{}}{ \hspace{-0.15cm}{6.21}\scalebox{0.99}\,($\pm$0.07)\hspace{-0.15cm} }\end{tabular}
\\ \midrule
\multirow{2}{*}{\hspace{-0.0cm}{\parbox{1.5cm}{\centering{\textbf{Uniformly Abrupt}}}}\hspace{-0.15cm}} 
& \multirow{1}{*}{\makecell[c]{VTAB-19T}}
& \begin{tabular}[c]{@{}c@{}}{ \hspace{-0.15cm}{6.15}\scalebox{0.99}\,($\pm$0.42)\hspace{-0.15cm} }\end{tabular}
& \begin{tabular}[c]{@{}c@{}}{ \hspace{-0.15cm}{6.43}\scalebox{0.99}\,($\pm$0.55)\hspace{-0.15cm} }\end{tabular}
& \begin{tabular}[c]{@{}c@{}}{ \hspace{-0.15cm}{9.48}\scalebox{0.99}\,($\pm$0.54)\hspace{-0.15cm} }\end{tabular}
& \begin{tabular}[c]{@{}c@{}}{ \hspace{-0.15cm}{5.50}\scalebox{0.99}\,($\pm$0.42)\hspace{-0.15cm} }\end{tabular}
& \begin{tabular}[c]{@{}c@{}}{ \hspace{-0.15cm}{{4.01}}\scalebox{0.99}\,($\pm$0.23)\hspace{-0.15cm} }\end{tabular}
& \begin{tabular}[c]{@{}c@{}}{ \hspace{-0.15cm}{{4.30}}\scalebox{0.99}\,($\pm$0.69)\hspace{-0.15cm} }\end{tabular}
\\ 
& \multirow{1}{*}{\makecell[c]{VTAB-5T}}
& \begin{tabular}[c]{@{}c@{}}{ \hspace{-0.15cm}{15.55}\scalebox{0.99}\,($\pm$1.13)\hspace{-0.15cm} }\end{tabular}
& \begin{tabular}[c]{@{}c@{}}{ \hspace{-0.15cm}{15.88}\scalebox{0.99}\,($\pm$1.09)\hspace{-0.15cm} }\end{tabular}
& \begin{tabular}[c]{@{}c@{}}{ \hspace{-0.15cm}{17.72}\scalebox{0.99}\,($\pm$1.29)\hspace{-0.15cm} }\end{tabular}
& \begin{tabular}[c]{@{}c@{}}{ \hspace{-0.15cm}{15.57}\scalebox{0.99}\,($\pm$1.07)\hspace{-0.15cm} }\end{tabular}
& \begin{tabular}[c]{@{}c@{}}{ \hspace{-0.15cm}{{12.84}}\scalebox{0.99}\,($\pm$0.98)\hspace{-0.15cm} }\end{tabular}
& \begin{tabular}[c]{@{}c@{}}{ \hspace{-0.15cm}{{13.25}}\scalebox{0.99}\,($\pm$1.00)\hspace{-0.15cm} }\end{tabular}
\\ \addlinespace[0.3ex]\cmidrule[1pt]{1-8}\addlinespace[0.1ex]
\multicolumn{2}{c|}{{ {\textbf{Avg. Forgetting\,(Degrad.)}}}} 
& \begin{tabular}[c]{@{}c@{}}{ \hspace{-0.15cm}{6.32}\scalebox{0.99}\,(-5.06\%)\hspace{-0.15cm} }\end{tabular}
& \begin{tabular}[c]{@{}c@{}}{ \hspace{-0.15cm}{6.65}\scalebox{0.99}\,(-9.77\%)\hspace{-0.15cm} }\end{tabular}
& \begin{tabular}[c]{@{}c@{}}{ \hspace{-0.15cm}{8.28}\scalebox{0.99}\,(-27.54\%)\hspace{-0.15cm} }\end{tabular}
& \begin{tabular}[c]{@{}c@{}}{ \hspace{-0.15cm}{6.26}\scalebox{0.99}\,(-4.15\%)\hspace{-0.15cm} }\end{tabular}
& \begin{tabular}[c]{@{}c@{}}{ \hspace{-0.15cm}{{6.49}}\scalebox{0.99}\,(-7.55\%)\hspace{-0.15cm} }\end{tabular}
& \begin{tabular}[c]{@{}c@{}}{ \hspace{-0.15cm}{{6.00}}\scalebox{0.99}\,(-\%)\hspace{-0.15cm} }\end{tabular}
\\ \bottomrule
\end{tabular}
}
\label{tbl:overall_fgt}
\end{table*}

To supplement the last accuracy $A_{last}$ presented in Table \ref{tbl:overall_perf}, we utilize the CL performance metric \emph{forgetting}, denoted by $F_{last}= \frac{1}{T-1}\Sigma_{j=1}^{T-1}f_{T,j}$. Here, $f_{i,j}$ indicates how much the model forgets about the $j$-th task after learning the $i$-th task\,($j<i$). It is important to note that $A_{last}$ encompasses the overall model's capability in both acquiring new skills\,(plasticity) and preserving knowledge of earlier tasks\,(forgetting) while $F_{last}$ only serves as a supplementary measure to the last accuracy $A_{last}$ while \,\cite{smith2023coda}.

Table \ref{tbl:overall_fgt} reveals that \algname{} achieves superior memory retention, exhibiting an average improvement in forgetting $F_{last}$ by 27.54\% and 7.55\% compared to the most accurate universal\,(LAE) and specific\,(S-Prompts) prompting methods across all scenarios, respectively. 
The success of \algname{} is primarily due to its adaptive prompt tuning approach, which tailors prompts to semantically similar tasks. This strategy effectively mitigates negative transfer between distinct tasks and curtails the forgetting of previously acquired knowledge during the acquisition of new information.






\section{Clustering Metrics}
\label{app_sec:clst_metrics}
To measure the correctness of semantic grouping by \algname{}, we adopt two widely-used clustering metrics: Adjusted Rand Index\,(ARI)\,\cite{steinley2004properties} and Normalized Mutual Information\,(NMI)\,\cite{hubert1985comparing}. They are commonly used to evaluate the performance of clustering algorithms by comparing their ability to correctly cluster data points to the labels of ground truth clustering.
In particular, ARI assesses the degree of agreement between pairs of samples by examining whether pairs from the same or different clusters remain consistent across the two clusterings. On the other hand, NMI evaluates the extent to which one clustering provides insights into the other, thereby determining the degree of mutual dependence between the cluster assignments. Specifically, the score of ARI ranges from -1 to 1 where a score of 1 indicates perfect agreement between two clusterings, 0 suggests no better than chance agreement, and negative values indicate disagreement. Likewise, NMI ranges from 0 to 1 where a score of 1 means perfect correlation\,(identical clusterings), and 0 indicates no mutual information\,(independent clusterings).

\section{Additional Visualization of Semantic Groups by \algname{}}
\label{app_sec:add_vis}

\begin{figure*}[ht!]
\begin{center}
\includegraphics[width=0.95\textwidth,]{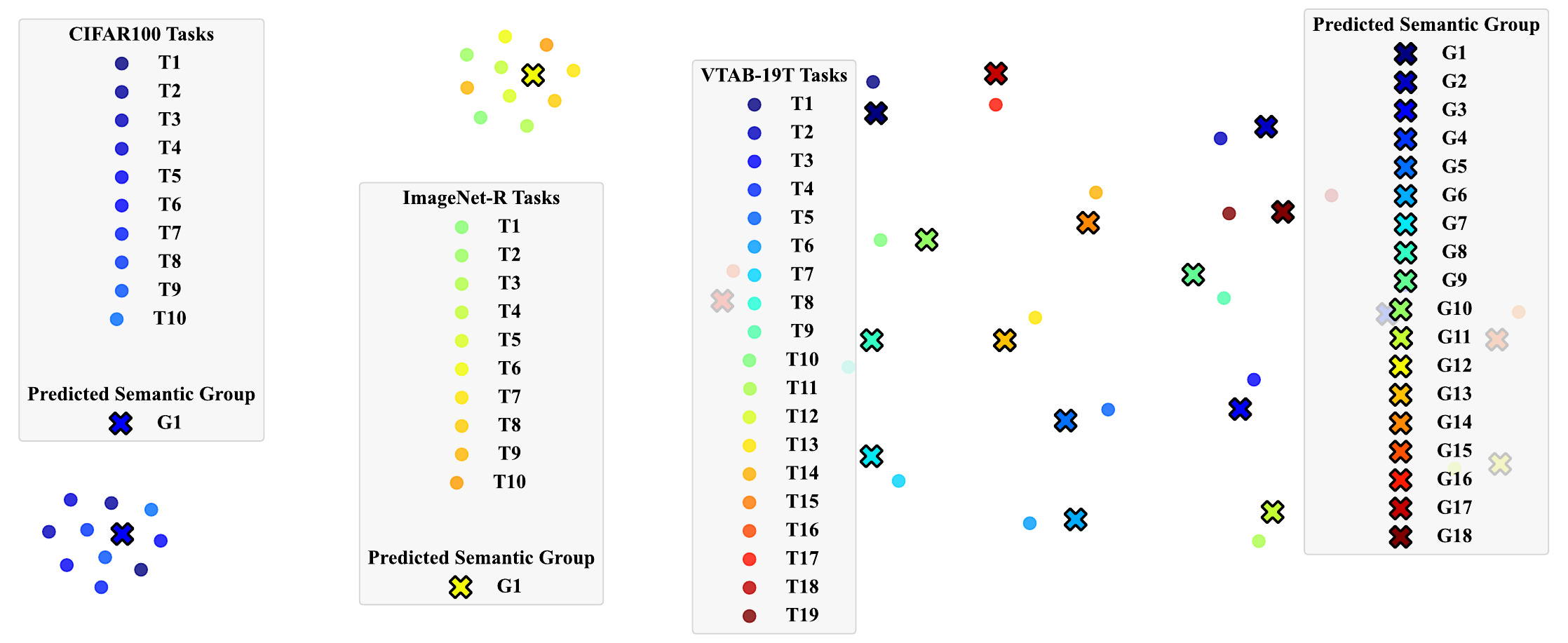}    
\end{center}
\hspace*{1.5cm} {\small (a) Uniformly mild shifting scenario.} \hspace*{3.cm} {\small (b) Uniformly
 abrupt shifting scenario.}
\vspace*{0.1cm}
\caption{t-SNE visualization the semantic groups formed by \algname{} for tasks from the uniformly mild shifting scenarios\,(CIFAR100, ImageNet-R), and abrupt shifting scenario\,(VTAB-19T). Tasks are denoted by circles \scalebox{0.99}{\color{gray}\CIRCLE}  and the symbols {\textbf{\texttimes{}}} represent the semantic groups identified by \algname{}. A single semantic group encompasses all tasks for CIFAR100 and ImageNet-R while distinct semantic groups are formed for nearly every task in the VTAB-19T dataset. }
\vspace*{0.1cm}
\label{fig:vis_groups_diverse_scenarios}
\end{figure*}

Figure \ref{fig:vis_groups_diverse_scenarios} complements Figure \ref{fig:vis_groups} by demonstrating the semantic groups formed by \algname{} under uniformly mild and abrupt shifting scenarios. It is evident that the emergence of these semantic groups is dependent on the degree of semantic shift. For cases with mild semantic shifts, such as CIFAR100 and ImageNet-R, all tasks are aggregated into a single semantic group, similar to the approach of universal prompting methods. Conversely, in the scenario with abrupt semantic shifts, the semantic groups tend to correspond nearly on a per-task basis, which is in line with specific prompting methods.

\section{Additional Parameter Sensitivity Analysis}
\begin{figure*}[ht!]
\begin{center}
\includegraphics[width=0.7\textwidth,]{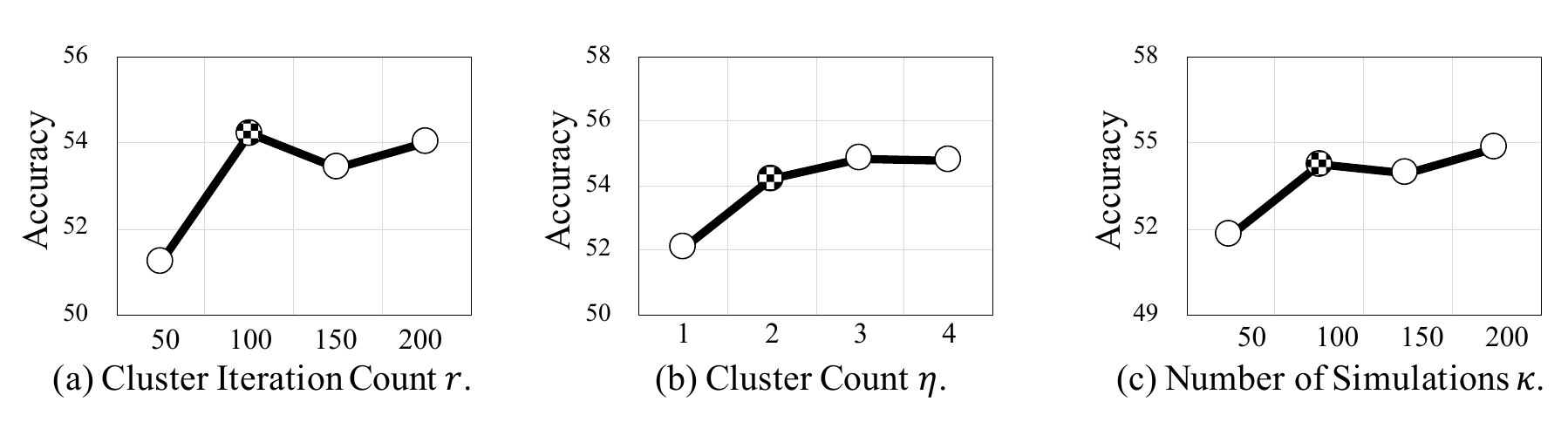}    
\end{center}
\vspace*{-0.3cm}
\caption{Parameter sensitivity analysis of clustering iteration count $r$, number of representative clusters $\eta$, and permutation sampling size $\kappa$ on prospective prompt collection and semantic refinement, using VTAB-Rec10. }
\label{fig:param_sensitivity_additional}
\end{figure*}

Figure \ref{fig:param_sensitivity_additional} presents the sensitivity analysis on the hyperparameters: the clustering iteration count $r$ and and the number of representative clusters $\eta$, as applied to the collection of prospective prompts in Eq.~\eqref{eq:Preparatory groups}, and the simulation sampling size $\kappa$, utilized for semantic refinement in Eq.~\eqref{eq:min_group}. 

\label{app_sec:add_param_sensitivity}
\noindent\textbf{Effect of Clustering Iteration Count.} 
Increasing clustering iteration count facilitates the creation of more comprehensive and diverse prospective semantic groups, which enhances the chances of refining to more generalizable semantic groups. Figure \ref{fig:param_sensitivity_additional}(a) illustrates the impact of different clustering iteration counts, specifically $r \in \{50, 100, 150, 200\}$, on test accuracy in varying shifting scenarios where the active refinement is required. Our findings show that there is no significant improvement in performance when the clustering iteration count exceeds 100. This suggests that a count of 100 is adequate for efficiently generating sufficiently comprehensive prospective semantic groups for semantic refinement. Consequently, we adopt $r= 100$ for all experiments.

\smallskip
\noindent\textbf{Effect of Representative Cluster Count.} 
A higher number of representative clusters can prepare more encompassing and varied prospective semantic groups, which in turn supports semantic refinement. Figure \ref{fig:param_sensitivity_additional}(b) presents the influence of varying representative cluster counts $\eta \in \{1, 2, 3, 4\}$ on test accuracy. We observe that a representative count of 2 or more achieves high accuracies, thereby suggesting that a count of 2 is adequate for the efficient formation of sufficiently comprehensive prospective semantic groups necessary for effective semantic refinement. Thus, for all experiments, we set the representative cluster count $\eta$ to 2. 

\smallskip
\noindent\textbf{Effect of Simulation Sampling Size.} 
The higher number of task order simulations increases the likelihood of identifying more generalizable semantic groups through refinement. Figure \ref{fig:param_sensitivity_additional}(c) shows the effect of sampling size $\kappa \in \{50, 100, 150, 200\}$ on the test accuracy in varying shift scenarios where refinement is actively required. 
Overall, our findings reveal that performance is not significantly affected by the number of simulations, particularly when the sampling size exceeds 100, indicating no meaningful improvement in performance. Accordingly, we establish $\kappa=100$ as the default sampling size for all experiments.



\end{document}